\documentclass[acmsmall]{acmart}
\usepackage[ruled, linesnumbered]{algorithm2e} 

\usepackage{threeparttable}
\usepackage{multirow}
\usepackage{subcaption}
\usepackage{amsmath}
\usepackage{relsize}
\DeclareMathOperator*{\argmin}{argmin}

\AtBeginDocument{%
  \providecommand\BibTeX{{%
    \normalfont B\kern-0.5em{\scshape i\kern-0.25em b}\kern-0.8em\TeX}}}

\setcopyright{acmcopyright}
\copyrightyear{2020}
\acmYear{2020}
\acmDOI{10.1145/1122445.xxxxxxxxxx}

\acmJournal{TALLIP}
\acmVolume{19}
\acmNumber{6}
\acmArticle{1}
\acmMonth{10}

\begin{document}

\title{Plan Optimization to Bilingual Dictionary Induction for Low-Resource Language Families}

\author{Arbi Haza Nasution}
\authornote{This is the corresponding author}
\orcid{0000-0001-6283-3217}
\affiliation{%
	\institution{Universitas Islam Riau}
	\department{Informatics Engineering}
	\city{Pekanbaru}
	\state{Riau}
	\country{Indonesia}
}
\email{arbi@eng.uir.ac.id}
\author{Yohei Murakami}
\affiliation{%
	\institution{Ritsumeikan University}
	\department{Faculty of Information Science and Engineering}
	\city{Kyoto}
	\country{Japan}
}
\email{yohei@fc.ritsumei.ac.jp}
\author{Toru Ishida}
\affiliation{%
	\institution{Waseda University}
	\department{School of Creative Science and Engineering}
	\city{Tokyo}
	\country{Japan}
}
\email{toru.ishida@aoni.waseda.jp}

\renewcommand{\shortauthors}{A. H. Nasution, et al.}

\begin{abstract}
  Creating bilingual dictionary is the first crucial step in enriching low-resource languages. Especially for the closely-related ones, it has been shown that the constraint-based approach is useful for inducing bilingual lexicons from two bilingual dictionaries via the pivot language. However, if there are no available machine-readable dictionaries as input, we need to consider manual creation by bilingual native speakers. To reach a goal of comprehensively create multiple bilingual dictionaries, even if we already have several existing machine-readable bilingual dictionaries, it is still difficult to determine the execution order of the constraint-based approach to reducing the total cost. Plan optimization is crucial in composing the order of bilingual dictionaries creation with the consideration of the methods and their costs. We formalize the plan optimization for creating bilingual dictionaries by utilizing Markov Decision Process (MDP) with the goal to get a more accurate estimation of the most feasible optimal plan with the least total cost before fully implementing the constraint-based bilingual lexicon induction. We model a prior beta distribution of bilingual lexicon induction precision with language similarity and polysemy of the topology as $\alpha$ and $\beta$ parameters. It is further used to model cost function and state transition probability. We estimated the cost of all investment plan as a baseline for evaluating the proposed MDP-based approach with total cost as an evaluation metric. After utilizing the posterior beta distribution in the first batch of experiments to construct the prior beta distribution in the second batch of experiments, the result shows 61.5\% of cost reduction compared to the estimated all investment plan and 39.4\% of cost reduction compared to the estimated MDP optimal plan. The MDP-based proposal outperformed the baseline on the total cost.
\end{abstract}

\ccsdesc[500]{Computing methodologies~Value iteration}
\ccsdesc[500]{Computing methodologies~Planning under uncertainty}
\ccsdesc[300]{Theory of computation~Constraint and logic programming}
\ccsdesc[500]{Computing methodologies~Language resources}
\ccsdesc[300]{Computing methodologies~Lexical semantics}
\ccsdesc[300]{Mathematics of computing~Bayesian networks}
\ccsdesc[300]{Mathematics of computing~Distribution functions}

\keywords{plan optimization, low-resource languages, closely-related languages, pivot-based bilingual lexicon induction}

\maketitle

\section{Introduction}
Machine-readable bilingual dictionaries are important language resources which are often utilized as language services \cite{ishida2018langrid} for various purpose such as supporting intercultural communication and collaboration \cite{ishida2016intercultural,nasution2017PHMT,nasution2018PHMT}. Unfortunately, low-resource languages lack such resources. Previous study on high-resource languages showed the effectiveness of parallel corpora \cite{Fung-98,brown1990statistical} and comparable corpora \cite{rapp1995identifying,fung1995compiling} in extracting bilingual lexicons. It is clear that bilingual lexicon extraction is not an easy task, yet challenging for low-resource languages due to the lack of parallel and comparable corpora. We introduced the promising approach of treating pivot-based bilingual lexicon induction for low-resource languages as an optimization problem \cite{nasution2017pivot} where the only language resources required as input are two bilingual dictionaries. In spite of the great potential of our constraint-based bilingual lexicon induction in enriching low-resource languages, when actually implementing the induction method, we need to consider adding a more traditional method to the equation, i.e., manually creating the bilingual dictionaries by bilingual native speakers. Despite the high cost, the inclusion of the manual creation will be unavoidable if no machine-readable dictionaries are available. When we want to comprehensively create all combination of bilingual dictionaries from a set of target languages, even if we already have several existing machine readable bilingual dictionaries, it is still difficult to determine the execution order of the constraint-based method to reducing the total cost. Moreover, when the constraint-based method failed to return the satisfiable size of output bilingual dictionary, the manual creation will fill in the gap. Considering the methods and their costs, we recently introduced a plan optimizer to find a feasible optimal plan of creating multiple bilingual dictionaries with the least total cost \cite{nasution2017plan}. The plan optimizer will calculate the best bilingual dictionary creation method (constraint-based induction or manual creation by human) to take in order to obtain all possible combination of bilingual dictionaries from the language set with the minimum total cost to be paid. However, the paper lacks actual data and experiment. It only presents a comparative simulation of the proposed MDP model and three heuristic models with an estimated total cost as a measure. The state transition probability modeling is also too naive as the precision of constraint-based bilingual lexicon induction assumed to be equals or exceeds input languages similarity. To obtain a better estimation of constraint-based bilingual lexicon induction precision and a better plan than our previous work, we extend the plan optimizer and address the following research goals:
\begin{itemize}
	\item{\textit{Modeling prior beta distribution of constraint-based bilingual lexicon induction precision}: We model language similarity and polysemy of the topology as beta distribution parameters.}
	\item{\textit{Formalization of plan optimization in creating bilingual dictionaries using Markov Decision Process}: Modeling bilingual dictionary dependency with AND/OR graphs as states, modeling constraint-based bilingual lexicon induction and manual dictionary creation by human as actions, and utilizing beta distribution of constraint-based bilingual lexicon induction precision to model cost function and state transition probability.}	
	\item{\textit{Evaluating the plan optimizer:} We evaluate the generated plan by conducting an experiment to create 10 bilingual dictionaries from 5 languages following the plan.}
\end{itemize}

The rest of this paper is organized as follows: We will briefly discuss a motivating scenario to lead reader into understanding the whole picture of our approach in Section 2. In Section 3, we will explain related research on pivot-based bilingual lexicon induction and introduce our novel modeling of constraint-based bilingual lexicon induction precision prior beta distribution. Section 4 provides details on how to model dictionary dependency. The plan optimization formalization, a core component of our proposal is discussed in Section 5. Section 6 describes our experiments and the results. Finally, Section 7 discuss the potential dynamic use of plan optimization and Section 8 concludes this paper.
\section{Motivating Scenario}
In order to illustrate the needs of optimal plan for creating multiple bilingual dictionaries with the least total cost we present an example motivating scenario. Consider a stakeholder has a motivation to obtain all 10 combination of bilingual dictionaries from 5 languages with a minimum size of 2,000 translation pairs each. Currently, the stakeholder already has a bilingual dictionary of language 1 and 3 ($d_{(1,3)}$) with 2,100 translation pairs and two bilingual dictionaries ($d_{(1,2)}$ and $d_{(2,3)}$) with a number of translation pairs below 2,000. Obviously, the stakeholder can just hire native speakers to create and evaluate the bilingual dictionaries following the traditional investment plan to reach his goal with a total cost of $C$. However, he can save cost of bilingual dictionary creation by utilizing our constraint-based bilingual lexicon induction with a zero creation cost. Even though the resulting bilingual dictionary still needs to be evaluated by native speakers, by following the optimal plan, the stakeholder can cut about half of the total cost.

At this point, the reader might wonder that even before executing the optimal plan, how can we know that utilizing the constraint-based bilingual lexicon induction to enrich $d_{(2,3)}$ resulting a satisfying size bilingual dictionary above 2,000 translation pairs or below 2,000 translation pairs that need to be invested more by native speakers to fill in the gap. To answer this question, the constraint-based bilingual lexicon induction precision need to be estimated in order to calculate the resulting size bilingual dictionary. This uncertainty is the research challenge that we want to address in the following sections by modeling beta distribution of constraint-based bilingual lexicon induction precision and further utilize it in formalizing plan optimization in creating bilingual dictionaries using Markov Decision Process (MDP), since MDP can handle planning under uncertainty. If one try to utilize both our constraint-based bilingual lexicon induction and manual creation by native speakers and try to create the plan (order of dictionary creation task to take) manually without our MDP approach, the total cost might be higher than following our MDP plan. Since the created bilingual dictionary can be used as input for inducing the other unsatisfying size dictionary, the order of dictionary creation task to take is crucial.
\section{Constraint-Based Bilingual Lexicon Induction}
\begin{figure*}[!t]
	\begin{center}
		\includegraphics[scale=0.6]{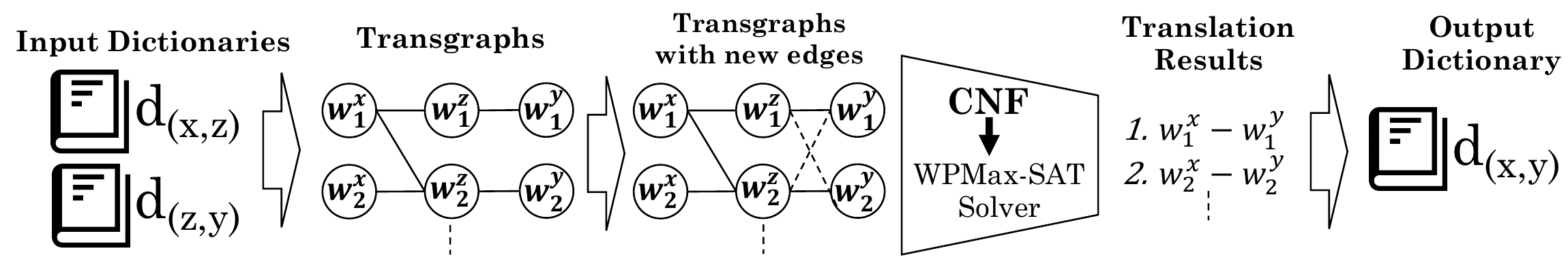} 
		\caption{One-to-one constraint approach to pivot-based bilingual lexicon induction.}
		\label{fig.1-1}
	\end{center}
\end{figure*}
The first work on bilingual lexicon induction to create bilingual dictionary of language x and language y, $d_{(x,y)}$, via pivot language z is Inverse Consultation (IC) \cite{tanaka-94}. It utilizes the structure of input dictionaries to measure the closeness of word meanings and then uses the results to trim incorrect translation pair candidates. The approach identifies equivalent candidates of language x words in language y by consulting $d_{(x,z)}$ and $d_{(z,y)}$. These equivalent candidates will be looked up and compared in the inverse dictionary $d_{(y,x)}$.

The pivot-based approach is very suitable for low-resource languages, especially when dictionaries are the only language resource required. Unfortunately, for some low-resource languages, it is often difficult to find machine-readable inverse dictionaries and corpora to identify and eliminate the incorrect translation pair candidates. To overcome this limitation, our team \cite{wushouer-15} proposed to treat pivot-based bilingual lexicon induction as an optimization problem. They assume that closely-related languages share a significant number of cognates (words with similar spelling/form and meaning originating from the same root language), thus one-to-one lexicon mapping should often be found. This assumption yielded the development of a constraint optimization model to induce an Uyghur-Kazakh bilingual dictionary using Chinese language as the pivot, which means that Chinese words were used as bridges to connect Uyghur words in an Uyghur-Chinese dictionary with Kazakh words in a Kazakh-Chinese dictionary. The proposal uses a graph whose vertices represent words and edges indicate shared meanings; following \cite{Soderland-09} it was called a transgraph. The proposal proceeds as follows.
\begin{enumerate}
	\item Use two bilingual dictionaries as input.
	\item Represent them as transgraphs where $w_1^x$ and $w_2^x$ are non-pivot words in language x, $w_1^z$ and $w_2^z$ are pivot words in language z, and $w_1^y$, $w_2^y$ and $w_3^y$ are non-pivot words in language y.
	\item Add some new edges represented by dashed edges based on the one-to-one assumption.
	\item Formalize the problem into conjunctive normal form (CNF) and use the Weighted Partial MaxSAT (WPMaxSAT) solver \cite{ansotegui2009solving} to return the optimized translation results.
	\item Output the induced bilingual dictionary as the result.
\end{enumerate}
These steps are shown in Figure~\ref{fig.1-1}. However, the assumption of one-to-one mapping is too strong to induce the many translation pairs needed to offset resource paucity because few such pairs can be found. Therefore, we generalized the constraint-based bilingual lexicon induction by extending constraints and translation pair candidates from the one-to-one approach to attain more voluminous bilingual dictionary results with many-to-many translation pairs extracted from connected existing and new edges \cite{nasution2016pivot}. We further enhance our generalized method by setting two steps to obtaining translation pair results. First, we identify one-to-one cognates by incorporating more constraints and heuristics to improve the quality of the translation result. We then identify the cognates' synonyms to obtain many-to-many translation pairs. In each step, we can obtain more cognate and cognate synonym pair candidates by iterating the n-cycle symmetry assumption until all possible translation pair candidates have been reached \cite{nasution2017pivot}.

\subsection{Modeling Prior Beta Distribution of Constraint-Based Bilingual Lexicon Induction Precision}
\begin{figure}[!t]
	\begin{center}
		\includegraphics[scale=0.6]{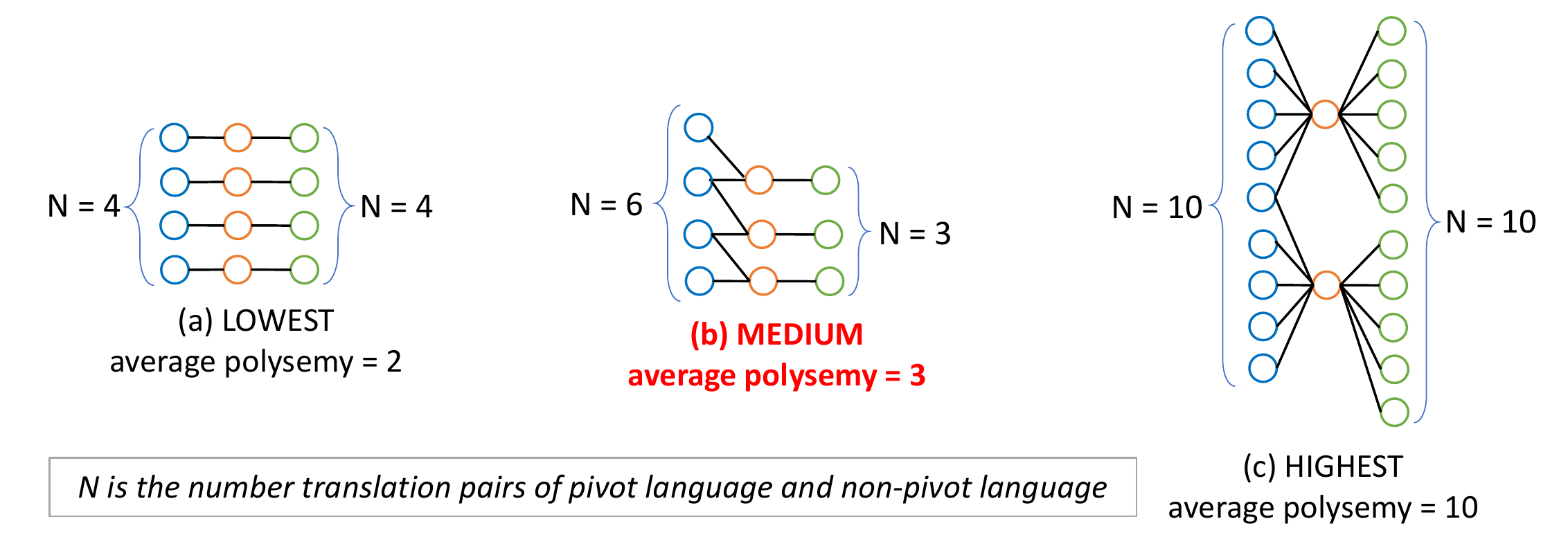}
		\caption{Average polysemy of the topology.}
		\label{fig. polysemy}
	\end{center}		
\end{figure}
The constraint-based bilingual lexicon induction has characteristics where it work better on closely-related languages and a higher polysemy pivot rate will hurt the precision. Having these positive and negative parameters, a beta distribution is the best distribution to model the constraint-based bilingual lexicon induction precision. A beta distribution is a family of continuous probability distributions defined on the interval [0, 1] parametrized by two positive shape parameters, denoted by $\alpha$ which positively affecting the probability (x-axis) and $\beta$ which negatively affecting the probability (x-axis). The two parameters control the shape of the distribution. 
Beta distribution is usually used in Bayesian statistics as prior distribution for either a proportion, or the probability of occurrence of an event, or the value of any random variable [0, 1] such as the reliability of a component \cite{gupta2004betadist}. The constraint-based bilingual lexicon induction precision is useful to estimate the resulting bilingual dictionary size. However, before actually implementing our constraint-based bilingual lexicon induction, it is difficult to precisely know the precision beforehand. We can treat the precision as a random variable [0, 1] that can be modeled with a beta distribution. When sample observations are not available, a beta distribution can be defined by using subjective information \cite{Fente1999betadist}. A precision of the constraint-based bilingual lexicon induction for closely-related low-resource languages is likely to fall in the middle area between 0 and 1, and the likelihood is getting slimmer as the precision close to 0 or 1, therefore, the precision is better modeled with a bell-shaped beta distribution with $\alpha \ge 2$ and $\beta \ge 2$.

After determining the shape of beta distribution, we further model the $\alpha$ and $\beta$ parameters for the prior beta distribution. Since $\alpha$ has a positive contribution to the precision, a language similarity of the target dictionary is a best fit because our constraint-based bilingual lexicon induction works better on a closely-related languages \cite{nasution2017pivot}. Automated Similarity Judgment Program (ASJP) was proposed by \cite{holman2011automated} with the main goal of developing a database of Swadesh lists \cite{swadesh1955towards} for all of the world's languages from which lexical similarity or lexical distance matrix between languages can be obtained by comparing the word lists. We utilize ASJP to select the target languages used in our case studies. We calculate language similarity between each language pair following our previous work \cite{nasution2019simcluster}.

On the other hand, polysemy of the pivot word could cause a mistranslation when we induce a translation pair candidate from the connected edge in the transgraph as shown in Figure~\ref{fig.1-1}. However, considering low-resource languages have limited resources, our constraint-based bilingual lexicon induction only consider input bilingual dictionary as list of translation pairs without any additional information like part-of-speech or sense information. Therefore, we assume that an edge in a transgraph represents distinct sense/meaning. We define a polysemy of the topology as an average number of connected edges to pivot word in all transgraphs.
\begin{figure}[!t]
	\begin{center}
		\includegraphics[scale=0.6]{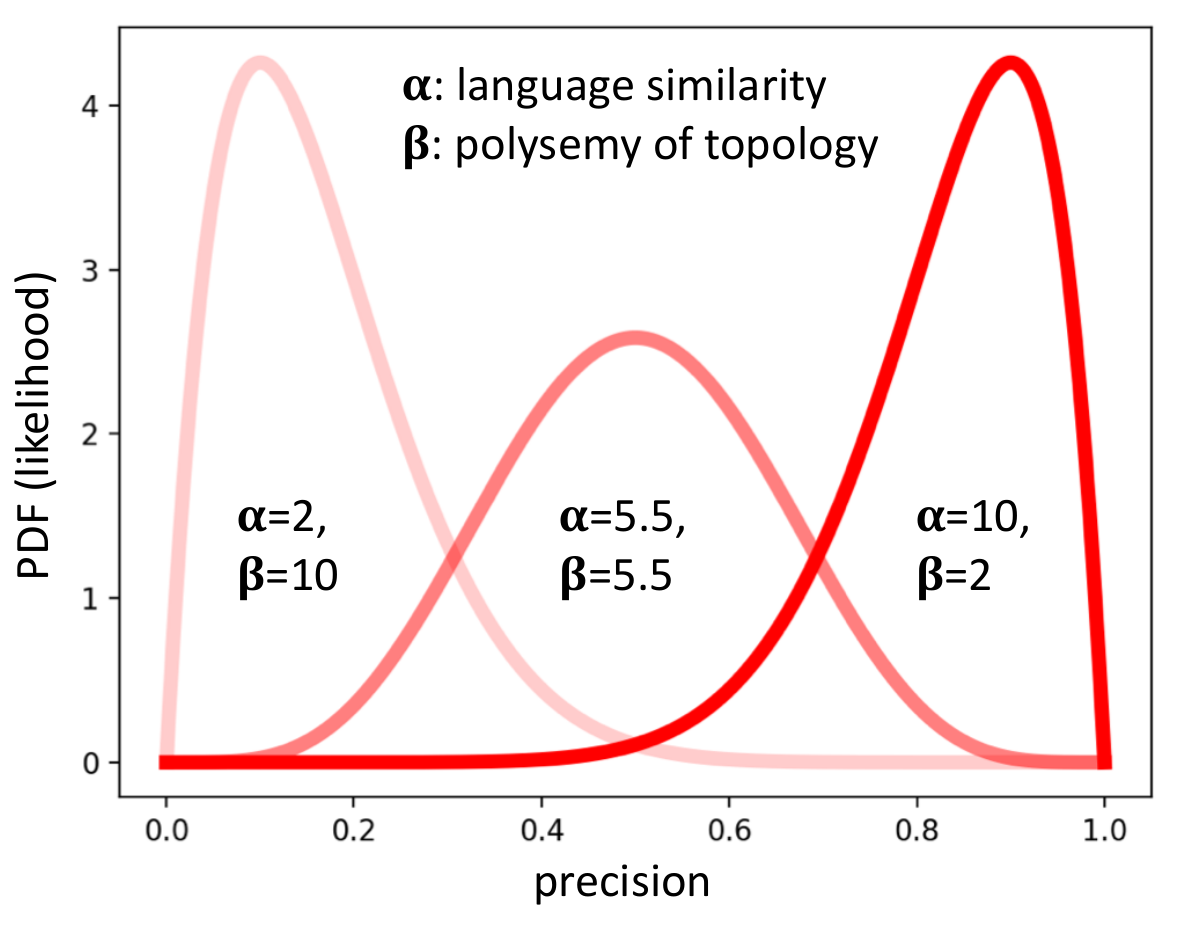}
		\caption{Variety of beta distribution bell-shaped depends on $\alpha$ and $\beta$.}
		\label{fig. beta}
	\end{center}		
\end{figure}
When a one-to-one topology rate is 1 which means that every pivot word is only connected to one word from each of the non-pivot language as shown in Figure~\ref{fig. polysemy}a, the polysemy of the topology is the lowest = 2. When each pivot word is connected to five word from each of the non-pivot language as shown in Figure~\ref{fig. polysemy}c, the polysemy of the topology is 10. We assume that the highest polysemy of topology is 10. The higher the polysemy of the topology, the more likely it is polysemous, hence negatively affect the constraint-based bilingual lexicon induction precision. So, we define $\beta$ as the polysemy of the topology ranging from 2 to 10. The language similarity is normalized into $\alpha \in [2, 10]$ to balance it with $\beta$. The beta distribution of constraint-based bilingual lexicon induction precision will have different bell-shaped depends on the $\alpha$ and $\beta$ parameters as shown in Figure~\ref{fig. beta}. The probability density function (PDF) is calculated by the following equation:
\begin{equation}
\label{eqn:PDF}
f(x;\alpha,\beta) =  \frac{1}{B(\alpha,\beta)} x^{\alpha-1}(1-x)^{\beta-1}; 0 < x < 1; \alpha,\beta \geq 2 
\end{equation}

\subsection{Modeling Dictionary Dependency} 
\begin{figure}[!t]
	\begin{center}
		\includegraphics[scale=0.5]{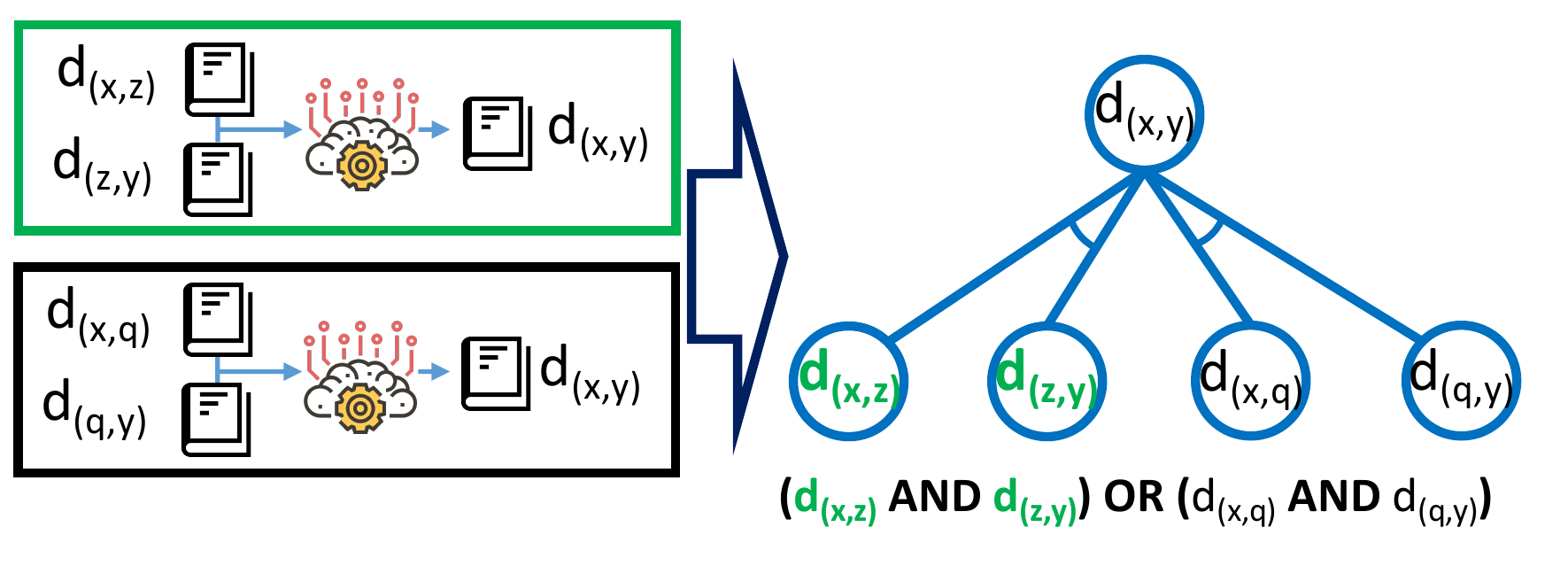}
		\caption{Modeling Bilingual Dictionary Induction Dependency.}
		\label{fig. dict_dependency}
	\end{center}		
\end{figure}

Our constraint-based bilingual lexicon induction requires two bilingual dictionaries that share the same pivot language. We can induce bilingual $d_{(x,y)}$ from $d_{(x,z)}$ and $d_{(z,y)}$ as input (language z is the pivot). Nevertheless, we can also induce $d_{(x,y)}$ with different input bilingual dictionaries using language q as the pivot language for instance. We use an AND/OR graph to model the dependency: bilingual $d_{(x,y)}$ can be induced from $d_{(x,z)}$ and $d_{(z,y)}$ OR from $d_{(x,q)}$ and $d_{(q,y)}$ as shown in Figure~\ref{fig. dict_dependency}.

If two sets of input dictionaries can be used to induce $d_{(x,y)}$, if we have to choose between the two sets, we need to prioritize input dictionaries that can induce $d_{(x,y)}$ with more correct translation pairs. But, the number of correct translation pairs that can be induced depends on the constraint-based bilingual lexicon induction precision and the size of translation pair candidates generated from the transgraph. 
\section{Formalizing Plan Optimization}
The plan optimization to bilingual lexicon induction involves discovering the order of bilingual dictionary creation task from a set of possible tasks including constraint-based bilingual lexicon induction and manual creation by native speakers to minimize the total cost. We assume that the number of existing translation pairs for existing bilingual dictionaries and the minimum number of translation pairs the output bilingual dictionary should have, $size(d^m_{(x,y)})$, are both known. Multiple candidate plans exist to finally obtain all bilingual dictionaries. One criteria for selecting a plan is to establish a model of optimality and select the plan that is most optimal. We formulate the plan optimization in the context of creating multiple bilingual dictionaries from a set of language of interest as a constraint optimization problem (CSP)~\cite{nasution2020towardsplanning}. Formally, a constraint satisfaction problem is defined as a triple  $\langle X,D,C\rangle$, where $X=\{X_{1},\ldots ,X_{n}\}$ is a set of variables, $D=\{D_{1},\ldots ,D_{n}\}$ is a set of the respective domains of values, and $C=\{C_{1},\ldots ,C_{m}\}$ is a set of constraints~\cite{russell2016artificial}.

\subsection{Variable}
If $n$ is a number of target languages specified, the total number of all possible combinations of target bilingual dictionaries is $h = {n\choose2} $. For example, if we have 4 languages ($L_1$, $L_2$, $L_3$, $L_4$), there will be $h = {4\choose2} = 6$ target bilingual dictionaries: $d_{(1,2)}$, $d_{(1,3)}$, $d_{(1,4)}$, $d_{(2,3)}$, $d_{(2,4)}$, and $d_{(3,4)}$. A state $S_i$ stores $h$ bilingual dictionaries, each $d_{(x,y)}$ has four possible status types:  not existing $d_{(x,y):n}$, existing but number of translation pairs is below minimum dictionary size requested by user: $d_{(x,y):eu}$, induced with constraint-based bilingual induction with $z$ as pivot language but the number of translation pairs is below minimum dictionary size requested by user: $d_{(x,y):pu(z)}$, and existing or manually created by native bilingual speakers or induced with constraint-based bilingual induction where the number of translation pairs equals or exceeds minimum dictionary size requested by user: $d_{(x,y):s}$, hence, the maximum number of state is $4^h= 4^6 = 4,096$. Based on the status, we further categorize the bilingual dictionary as either $SATDict$ ($d_{(x,y):s}$) or $UnSATDict$ ($d_{(x,y):n}$, $d_{(x,y):eu}$, or $d_{(x,y):pu(z)}$). 
A variable $X_i$ is a possible bilingual dictionary creation method applied to enrich the size hence changing the status of bilingual dictionaries inside state $S_i$. The number of state increases exponentially with the number of target languages. So as to cast formulation complexity into a graph theory problem, we initially create only one start state $S_1$ along with variable $X_1$ where each bilingual dictionary status is labeled based on the size of existing bilingual dictionaries given by user. The following states $S_2$, $S_3$, ..., $S_m$ and the respective variables $X_2$, $X_3$, ..., $X_m$ are created as each value in domain $D_i$ is defined.

\subsection{Domain}
Some bilingual dictionary creation methods such as the inverse consultation method, the one-to-one constraint-based approach, and our constraint-based bilingual lexicon induction require only bilingual dictionaries as input. However, since our method outperformed both previous methods, we model our method as one of value that can be assigned to variable $X_i$ and call it pivot action $a^p_{(x,z,y)}$ to create dictionary $d_{(x,y)}$ where $z$ is the pivot language. For low-resource languages, adequate machine-readable bilingual dictionaries are often unavailable, so, we define another value, manual bilingual dictionary creation by a native speaker as investment action $a^i_{(x,y)}$. The purposes of assigning the two values, the pivot action and investment action, are to enrich the size and change the category of the bilingual dictionaries stored in each state $S_i$ from $UnSATDict$ to $SATDict$. 

\subsection{Constraints for Domain Reduction}
\label{sec:constraintslist}
The following constraints are used to reduce the domain of a variable $X_i$.
\subsubsection{Adequate Dictionary Size Constraint ($C_1$)}
A dictionary $d_{(x,y)}$ inside a state $S_i$ cannot be created or enriched if the dictionary status is $d_{(x,y):s}$ where the number of translation pairs equals or exceeds minimum dictionary size requested by user, $size(d^m_{(x,y)})$. In other word, neither $a^i_{(x,y)}$ nor $a^p_{(x,z,y)}$; for any pivot language $z$  can be assigned to the variable $X_i$. If all dictionaries in a state $S_i$ have a status of $d_{(x,y):s}$, there are no available value to be assigned to variable $X_i$ in the domain $D_i$.
\subsubsection{Initial Dictionary Status Constraint ($C_2$)}
Initially, user provides information about the size of machine readable bilingual dictionaries if exist. The dictionary size information is mapped to a dictionary status of either $d_{(x,y):n}$, $d_{(x,y):eu}$, or $d_{(x,y):s}$. An $UnSATDict$ with status of $d_{(x,y):n}$ or $d_{(x,y):eu}$ inside a variable $X_i$ can be enriched by both investment action $a^i_{(x,y)}$ and pivot action $a^p_{(x,z,y)}$. Both values can be assigned to the variable $X_i$.
\subsubsection{One-Time Induction Constraint ($C_3$)}
For an $UnSATDict$ with status $d_{(x,y):pu(z)}$ inside a variable $X_i$, however, the next action is limited to investment action $a^i_{(x,y)}$ only, because pivot action $a^p_{(x,z,y)}$ was already executed exactly one step prior. Thus, investment action $a^i_{(x,y)}$ is the only possible value to be assigned to the variable $X_i$.
\subsubsection{Dictionary Induction Dependency Constraint ($C_4$)}
\begin{figure}[!t]
	\begin{center}
		\includegraphics[scale=0.56]{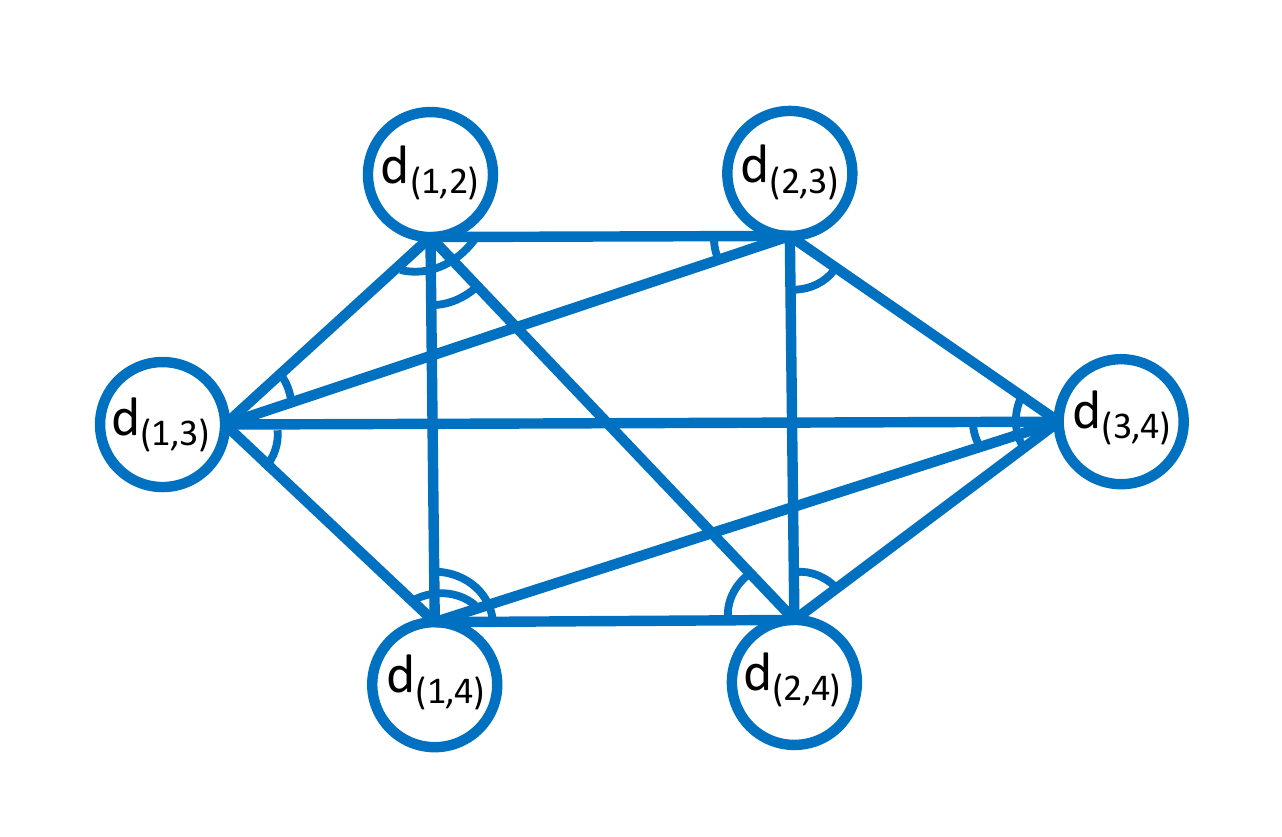} 
		\caption{Bilingual Dictionary Induction Dependency Model.}
		\label{fig.dependency model}
	\end{center}		
\end{figure}

A pivot action can be taken with a pair of input dictionary $d_{(x,z)}$ and $d_{(z,y)}$ as input when both of dictionaries have a status of $s$, where the number of translation pairs equals or exceeds minimum dictionary size requested by user, $eu$, which exists but the number of translation pairs is below minimum dictionary size requested by user, or  $pu(z)$ induced with constraint-based bilingual induction with $z$ as pivot language but the number of translation pairs is below minimum dictionary size requested by user. However, allowing dictionary with a status of $pu(z)$ as input can cause inconsistency of the translation pair result size. We consider the worst case scenario and choose the minimum translation pair result size. The bilingual lexicon induction dependency is shown in Figure~\ref{fig.dependency model}.

\subsection{Objective Function} 
In order to create or enrich bilingual dictionaries inside a state $S_i$, a constraint-based bilingual lexicon induction as pivot action $a^p_{(x,z,y)}$ or a manual bilingual dictionary creation by a native speaker as investment action $a^i_{(x,y)}$ can be assigned to a variable $X_i$. When we take an investment action, we are actually asking a native speaker to manually create and evaluate a bilingual dictionary and we need to pay for the time and effort incurred. On the other hand, for taking pivot action, i.e., using the constraint-based bilingual lexicon induction, when we already have the input dictionaries, we can generate the output dictionary in a short time. Thus, we assume that there is no cost for creating the bilingual dictionary, however, we still need to pay the native speaker to evaluate it.

Let $W$ be the set of candidate plans. Let $C(w, X_i, a)$ be the cost function associated with assigning a value $a$ in a corresponding domain $D_i$ for variable $X_i$ in some plan, $w$. The objective function is to minimize the expected total cost of assigning values in the corresponding domain to all variables while satisfying all four constraints. A plan optimization is a way to find an optimal plan, $w^*$, that results in the minimal expected total cost of assignment. Formally,
\begin{equation}
\label{eqn:objective function}
\begin{aligned}
&w^* = \argmin_{w \in W}
&&E \big( \sum_{a\in D_i}C(w, X_i, a)\big)\\
&\text{subject to}
&&\text{satisfying all constraints } {C_1, C_2, C_3, C_4}
\end{aligned}
\end{equation}		

The expectation operator, $E(.)$, in the above equation is necessary due to the stochastic nature of constraint-based bilingual lexicon induction. Based on the constraint-based bilingual lexicon induction precision, the resulting bilingual dictionary size can be above or below the minimum dictionary size requested by user, $size(d^m_{(x,y)})$. That is why we can only estimate the total cost before actually execute the task. A stochastic nature of the constraint-based bilingual lexicon induction is best handled by a Markov Decision Process (MDP), a well-known technique to solve problems containing uncertainty. Therefore, we model the plan application to bilingual dictionaries creation as a directed acyclic graph with MDP. A MDP has been used to model workflow composition and optimization \cite{doshi2004workflowcomposition,yu2005workflowgrid}. 
\begin{algorithm}[!t]
	\SetAlgoNoLine
	\footnotesize
	\DontPrintSemicolon
	\KwIn{
		targetLanguages, targetLanguageInfo, existingDictionaries\\
		\tcc{5 targetLanguages: [Indonesia "$ind$",Malay "$zlm$",Minangkabau "$min$",Javanese "$jav$",Sundanese "$sun$"]
			\\targetLanguageInfo is a list of pair of language similarities and $size(d^m_{(x,y)})=2,000$ 
			\\existingDictionaries=[$size(d_{(ind,zlm)})=711, size(d_{(ind,min)})=2,590, size(d_{(zlm,min)})=1,246$]}
	} 
	\KwOut{S, A, TS, T, C, dictionaryList \tcc*{Abbr: States, Actions, Target States, State Transition Probabilities, Costs}} 
	\tcc{Generate all ${5\choose2}=10$ combinations. Initialize the size to $0$ and status to not existing ($n$)}
	dictionaryList $\leftarrow$ generateDictionaryList(targetLanguages);\;
	\For{each $d_{(x,y)}$ in existingDictionaries}{
		dictionaryList.updateSizeAndStatus($d_{(x,y)}$);		
	}
	S[0] $\leftarrow$ createStartState(dictionaryList);\tcc*{In this example S[0] = [$d_{(ind,zlm):eu}$, $d_{(ind,min):s}$, $d_{(ind,jav):n}$, $d_{(ind,sun):n}$, $d_{(zlm,min):eu}$, $d_{(zlm,jav):n}$, $d_{(zlm,sun):n}$, $d_{(min,jav):n}$, $d_{(min,sun):n}$, $d_{(jav,sun):n}$]}
	unvisitedStates.add(S[0]);\;
	\While{unvisitedStates is not empty}{
		state $\leftarrow$ getStateWithLowestId(unvisitedStates);\;
		A[state] $\leftarrow$ createPossibleActions(state);\tcc*{Adhere to all constraints in Section~\ref{sec:constraintslist}}
		\For{each action in A[state]}{
			TS[state, action] $\leftarrow$ createTargetStates(state, action);\;
			\For{each targetState in TS[state, action]}{								
				T[state, action, targetState] $\leftarrow$ calculateTransitionProb(state, action, targetState, targetLanguageInfo);\tcc*{Section~\ref{sec:state transition}}
				C[state, action, targetState] $\leftarrow$ calculateCost(state, action, targetState, targetLanguageInfo);\tcc*{Section~\ref{sec:cost}}
				unvisitedStates.add(targetState);
			}
		}	
		unvisitedStates.remove(state);
	}	
	return S, A, TS, T, C, dictionaryList;
	\caption{State Transition Graph Generation} 
	\label{alg:formalization}
\end{algorithm}
\subsection{Markov Decision Process (MDP)} 
A MDP is a discrete time stochastic control process which provides a mathematical framework for modeling decision making in situations where outcomes are partly random and partly under the control of a decision maker. A MDP is often used for studying optimization problems solved via dynamic programming (value-iteration or policy-iteration) or reinforcement learning (Q-learning). Both value-iteration and policy-iteration assume that the agent knows the MDP model of the world (i.e. state-transition probability and reward/cost functions), in contrary, Q-learning does not know the model, it tries to learn the environment. In this paper, we use value-iteration method to find optimal policy for every state since we can estimate the state transition probability and the cost functions. A MDP is the tuple ($S$, $A$, $T(s,a,s')$, $C(s,a,s'))$, where $S$ is a set of states, $A$ is a set of actions, $T(s,a,s')$ is a transition probability distribution over the state space when action a is taken in state s, and $C(s,a,s')$ is the negative reward or cost for taking action a in state s. The formalization to MDP is described in Algorithm~\ref{alg:formalization}.
\subsubsection{State}
We model a MDP state similar with the way we define CSP variable. If $n$ is a number of target languages specified, the total number of all possible combinations of bilingual dictionaries in the state is $h = {n\choose2} $ as shown in Algorithm~\ref{alg:formalization} line number 1. Each state stores $h$ bilingual dictionaries, each $d_{(x,y)}$ with four possible status types:  not existing $d_{(x,y):n}$, existing but number of translation pairs is below minimum dictionary size requested by user: $d_{(x,y):eu}$, induced from pivot action with $z$ as pivot language but the number of translation pairs is below minimum dictionary size requested by user: $d_{(x,y):pu(z)}$, and existing or manually created by native bilingual speakers or induced with pivot action where the number of translation pairs equals or exceeds minimum dictionary size requested by user: $d_{(x,y):s}$, hence, the maximum number of MDP states is also $4^h= 4^6 = 4,096$. Based on the status, we further categorize the bilingual dictionary as either $SATDict$ ($d_{(x,y):s}$) or $UnSATDict$ ($d_{(x,y):n}$, $d_{(x,y):eu}$, or $d_{(x,y):pu(z)}$). 
After an agent takes an action in state $s$ to enrich an $UnSATDict$ of language $x$ and $y$, if the size of the output dictionary satisfies minimum dictionary size requested by user, the agent will transit to the next one step ahead state, $s'_{sat}$, which has an $SATDict$ of the same languages, $x$ and $y$, while the other bilingual dictionaries in $s'_{sat}$ are unchanged from the previous state, $s$. On the other hand, if the size of the output dictionary below user request, the agent will transit to the next one step ahead state, $s'_{unsat}$, which has an $UnSATDict$ of the same languages, $x$ and $y$, while the other bilingual dictionaries in $s'_{unsat}$ are unchanged from the previous state, $s$. 

The number of states increases exponentially with the number of languages. So as to cast formulation complexity into a graph theory problem, we initially create only one start state where each bilingual dictionary status is calculated based on the input bilingual dictionaries size given by user as shown in Algorithm~\ref{alg:formalization} line number 5. A list of unvisited states, $unvisitedStates$ is initialized with the start state as shown in line number 6. For each possible action of each state, target states are generated. Each target state which is not in $unvisitedStates$ list will be registered. After assigning all possible actions to the current state, it will be unregistered from the $unvisitedStates$ list as shown in line number 18. The iteration is stopped when the $unvisitedStates$ list is empty and the final state is reached where all $m$ bilingual dictionaries from $n$ languages are available and the number of translation pairs equals or exceeds user requested number of translation pairs.

\subsubsection{Action}
We also model a MDP action similar with the way we define CSP value in a domain. We apply our method as one of MDP action and call it pivot action $a^p_{(x,z,y)}$ to create dictionary $d_{(x,y)}$ where $z$ is the pivot language. We also define manual bilingual dictionary creation by a native speaker as investment action $a^i_{(x,y)}$. The purposes of the pivot action and investment action are to enrich and change the category of the bilingual dictionaries stored in each state from UnSATDict to SATDict. Adhering to CSP constraints, we assign all possible actions to a state based on the state's situation as shown in Algorithm~\ref{alg:formalization} line number 9. An UnSATDict with status $d_{(x,y):n}$ or $d_{(x,y):eu}$ can be enriched by both investment action and pivot action. For an UnSATDict with status $d_{(x,y):pu(z)}$, we limit the next action to investment action only because pivot action $a^p_{(x,z,y)}$ was already tried exactly one step prior. If other pivot such as $v$ is used to enrich  $d_{(x,y):pu(z)}$, there will be a redundancy issue on the output dictionary. We can not estimate the duplicate entries when we merge $d_{(x,y):pu(z)}$ and $d_{(x,y):pu(v)}$, thus, the output dictionary size will be misleading. A pivot action can be taken from input dictionaries with status $d_{(x,y):s}$, $d_{(x,y):eu}$, and $d_{(x,y):pu}$.

\begin{figure}[!t]
	\begin{center}
		\includegraphics[scale=0.7]{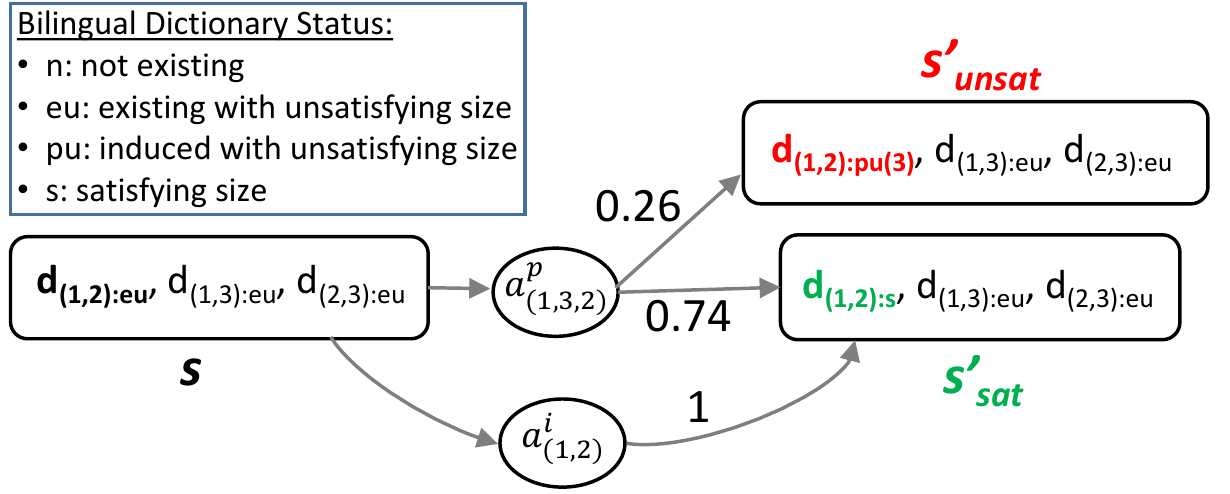} 
		\caption{Example of State Transition.}
		\label{fig.transition}
	\end{center}		
\end{figure}
\subsubsection{State Transition Probability}
\label{sec:state transition}
The state transition probability from a state $s$ to a target state $s'$ after taking an action is calculated as shown in Algorithm~\ref{alg:formalization} line number 13. The size of dictionaries in the current state affects performance of the pivot action taken in the current state, and thus the number of induced translation pairs in the next state. When the bilingual dictionary output by the pivot action $a^p_{(x,z,y)}$ in the current state $s$ equals or exceeds minimum dictionary size requested by user, the agent will transit to the next state, $s'_{sat}$ in which the bilingual dictionary status of languages $x$ and $y$ is $d_{(x,y):s}$ or else transit to the next state, $s'_{unsat}$ in which the bilingual dictionary status of languages $x$ and $y$ is $d_{(x,y):pu(z)}$ and the remaining bilingual dictionaries in the next state are unchanged from the previous state $s$ as shown in Figure~\ref{fig.transition}. In practice, we predict that the topology in Figure~\ref{fig. polysemy}b is more likely to be generated, so, we estimate the number of translation pair candidates, $size(d^{c}_{(x,y)})$, twice the minimum size of the two input dictionaries. Formally,
\begin{equation}
\label{eqn:translation candidate}
size(d^{c}_{(x,y)}) = 2 \times \min \big\{size(d_{(x,z)}), size(d_{(y,z)})\big\}
\end{equation}
The number of induced translation pairs is calculated by multiplying the pivot action precision with the number of translation pair candidates. Formally,
\begin{equation}
\label{eqn:induction size}
size(d_{(x,y)}) =  precision(a^p_{(x,z,y)}) \times size(d^{c}_{(x,y)})
\end{equation}
To calculate the required number of translation pairs to be induced or invested, for dictionary with the following status: $d_{(x,y):eu}$ or $d_{(x,y):pu(z)}$, it can be obtained by subtracting the minimum dictionary size requested by user to the dictionary size $size(d_{(x,y):eu})$ or $size(d_{(x,y):pu(z)})$. Formally,
\begin{equation}
\label{eqn:required size exist}
size(d^r_{(x,y)}) =  size(d^m_{(x,y)}) - size(d_{((x,y))})
\end{equation}
However, for empty dictionary with no existing translation pairs: $d_{(x,y):n}$, the required number of translation pairs to be induced or invested equals the minimum dictionary size requested by user. Formally,
\begin{equation}
\label{eqn:required size not exist}
size(d^r_{(x,y)}) =  size(d^m_{(x,y)})
\end{equation}
In order for $d_{(x,y)}$ to satisfy the required number of translation pairs, $size(d^r_{(x,y)})$, the pivot action precision should be at least equals to,
\begin{equation}
\label{eqn:min precision}
k =  \frac{size(d^r_{(x,y)})}{size(d^{c}_{(x,y)})}
\end{equation}
The state transition probability for taking a pivot action depends on the size of output bilingual dictionary which also depends on the precision of the constraint-based bilingual lexicon induction. If the precision is 1, then all translation pair candidates are taken as translation pairs. We model the state transition probability for taking a pivot action from the current state $s$ and fail to satisfy the minimum dictionary size requested by user, $size(d^r_{(x,y)})$ and going to $s'_{unsat}$ using beta distribution cumulative distribution function (CDF) ranging from 0 to $k$. Formally,
\begin{equation}
\label{eqn:CDF}
T(s,a,s'_{unsat}) = F(k;\alpha,\beta) =  \int_{0}^{k} f(x;\alpha,\beta) dx
\end{equation}
In the case of successfully satisfying the minimum dictionary size requested by user, $size(d^r_{(x,y)})$ and going to $s'_{sat}$, we use survival function. Formally,
\begin{equation}
\label{eqn:survival function}
T(s,a,s'_{sat}) = 1 - F(k;\alpha,\beta) = 1 - \int_{0}^{k} f(x;\alpha,\beta) dx
\end{equation}
\begin{figure}[!t]
	\begin{center}
		\includegraphics[scale=0.56]{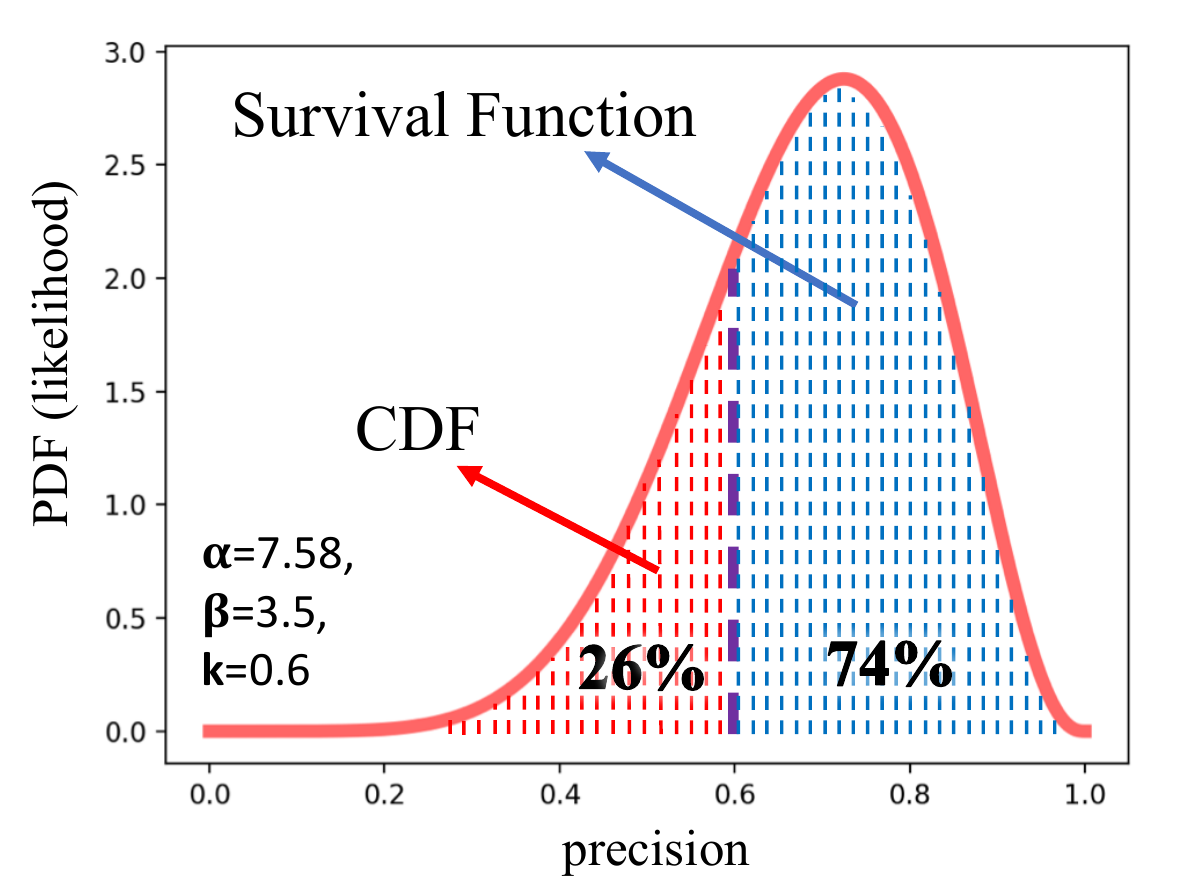} 
		\caption{Cumulative distribution function (CDF) and survival function.}
		\label{fig.CDF}
	\end{center}		
\end{figure}
For instance, when we want to enrich UnSATDict $d_{(1,2):eu}$ from an existing dictionary size of $4,000$ to a minimum dictionary size requested by user, $size(d^m_{(x,y)})=10,000$, we can calculate the required number of translation pairs to be induced with Equation~\eqref{eqn:required size exist}, $size(d^r_{(x,y)})=10,000-4,000=6,000$. If we enrich $d_{(1,2):eu}$ with pivot action $a^p_{(1,3,2)}$ from existing UnSATDict $d_{(1,3):eu}$ with input dictionary size equals $5,000$ and $d_{(2,3):eu}$ with input dictionary size equals $6,500$, using Equation~\eqref{eqn:translation candidate} we can get the number of translation pair candidates $size(d^{c}_{(1,2)})=2\times5,000=10,000$. Using Equation~\eqref{eqn:min precision}, we can calculate the minimum constraint-based bilingual lexicon induction precision, $k = 6,000 / 10,000 = 0.6$. If the beta distribution parameters are known, $\alpha=7.58, \beta=3.5$, using Equation~\eqref{eqn:CDF}, we can calculate the $T(s,a,s'_{unsat})=0.259$, and using Equation~\eqref{eqn:survival function}, we can calculate the $T(s,a,s'_{sat})=0.741$. As shown in Figure~\ref{fig.CDF}, there is 74\% probability of getting precision above the minimum constraint-based bilingual lexicon induction precision to satisfy the required number of translation pairs to be induced, thus agent will transit to $s'_{sat}$ and there is 26\% probability of getting precision below the minimum constraint-based bilingual lexicon induction precision to satisfy the required number of translation pairs to be induced, thus agent will transit to $s'_{unsat}$ as shown in Figure~\ref{fig.transition}.

\subsubsection{Cost}
\label{sec:cost}
In the  MDP model, the agent expects to get a reward after taking some actions. The reward will guide the agent to reach the final state and obtain the best path or in this case the best plan. Because for creating a bilingual dictionary we need to pay some cost instead of getting some rewards afterward, here we cast the reward as a cost. The terms of reward and cost are interchangeable in many previous MDP studies \cite{white1993survey}. The cost of taking an action $a$ from a state $s$ to a target state $s'$ is calculated as shown in Algorithm~\ref{alg:formalization} line number 14. When we take an investment action, we are actually asking a native speaker to manually create and evaluate a bilingual dictionary and we need to pay for the time and effort incurred, however, in the MDP model, we define the cost as duration/time taken to do the task. To calculate the cost of taking investment action $a\in A^i$ from state $s$ to state $s'$, the required number of translation pairs is multiplied by both $creationCost$ and $evaluationCost$. By estimating 0.8 human accuracy for manual dictionary creation, the cost of investment action is as follow,
\begin{equation}
\label{eqn:investment cost}
C(s,a,s') = \frac{size(d^r_{(x,y)})}{0.8} \times (creationCost + evaluationCost); a\in A^i
\end{equation}
On the other hand, for taking pivot action, i.e., using the constraint-based bilingual lexicon induction, when we already have the input dictionaries, we can generate the output dictionary in a short time. Thus, we assume that there is no cost for creating the bilingual dictionary, in other word, the $creationCost=0$, however, we still need to pay native speaker to evaluate it. To calculate the cost of taking pivot action $a\in A^p$ from state $s$ to state $s'$, the number of translation pair candidates is multiplied by $evaluationCost$.
\begin{equation}
\label{eqn:cost function}
C(s,a,s') = size(d^c_{(x,y)}) \times evaluationCost; a\in A^p
\end{equation}
Since the action cost, $C(s,a,s')$, for pivot action, depends on the number of translation pair candidates, $size(d^c_{(x,y)})$, and for investment action, depends on the required number of translation pairs, $size(d^r_{(x,y)})$, which are both calculated based on the size of the input dictionaries, which are unknown except for the existing dictionaries, we need to estimate the size of each dictionary in every state beforehand. This involves estimating the size of output dictionary in state $s'$ after taking investment action and pivot action in state $s$. Based on Equation~\eqref{eqn:investment cost}, estimating 0.8 human accuracy, we can easily predict the output dictionary by dividing the required number of translation pairs with 0.8, $size(d^r_{(x,y)}) / 0.8$. However, for pivot action, we need to estimate the precision of the constraint-based bilingual lexicon induction when the agent transit to $s'_{sat}$ and $s'_{unsat}$. To calculate the expected value (mean) of a beta distribution, we can use the following Equation:
\begin{equation}
\label{eqn:mean}
E(X) = \int_{0}^{1} xf(x;\alpha,\beta) dx = \frac{\alpha}{\alpha+\beta}
\end{equation}
However, the above equation consider the whole beta distribution, while we need to calculate upper mean and lower mean to estimate the precision of the constraint-based bilingual lexicon induction when the agent transit to $s'_{sat}$ and $s'_{unsat}$, respectively. To do this, firstly, we need to truncate the beta distribution of constraint-based bilingual lexicon induction precision by $k$, the minimum precision to satisfy minimum dictionary size requested by user, $size(d^m_{(x,y)})$, and further calculate the upper mean and lower mean of the truncated beta distribution. This mean of a truncated distribution is pretty straightforward with a beta. For a positive random variable we have
\begin{equation}
\label{eqn:truncated beta}
E(X|X < k) = \frac{\int_{0}^{k} x f(x;\alpha,\beta) dx}{\int_{0}^{k} f(x;\alpha,\beta) dx} 
\end{equation}
Moving from Equation~\eqref{eqn:PDF}, we have
\begin{equation}
\label{eqn:x f(x)}
x f(x;\alpha,\beta) =  \frac{B(\alpha+1,\beta)}{B(\alpha,\beta)} f(x;\alpha+1,\beta) = \frac{\alpha}{\alpha+\beta} f(x;\alpha+1,\beta)
\end{equation}
Substituting Equation~\eqref{eqn:x f(x)} to Equation~\eqref{eqn:truncated beta}, the mean of the truncated beta distribution is simplified as Equation~\eqref{eqn:lower mean} to calculate the lower mean of the truncated beta distribution to estimate the precision of the constraint-based bilingual lexicon induction when the agent transit to $s'_{unsat}$. Now the two integrals are just beta CDFs which are easily computed.
\begin{figure}[!t]
	\begin{center}
		\includegraphics[scale=0.56]{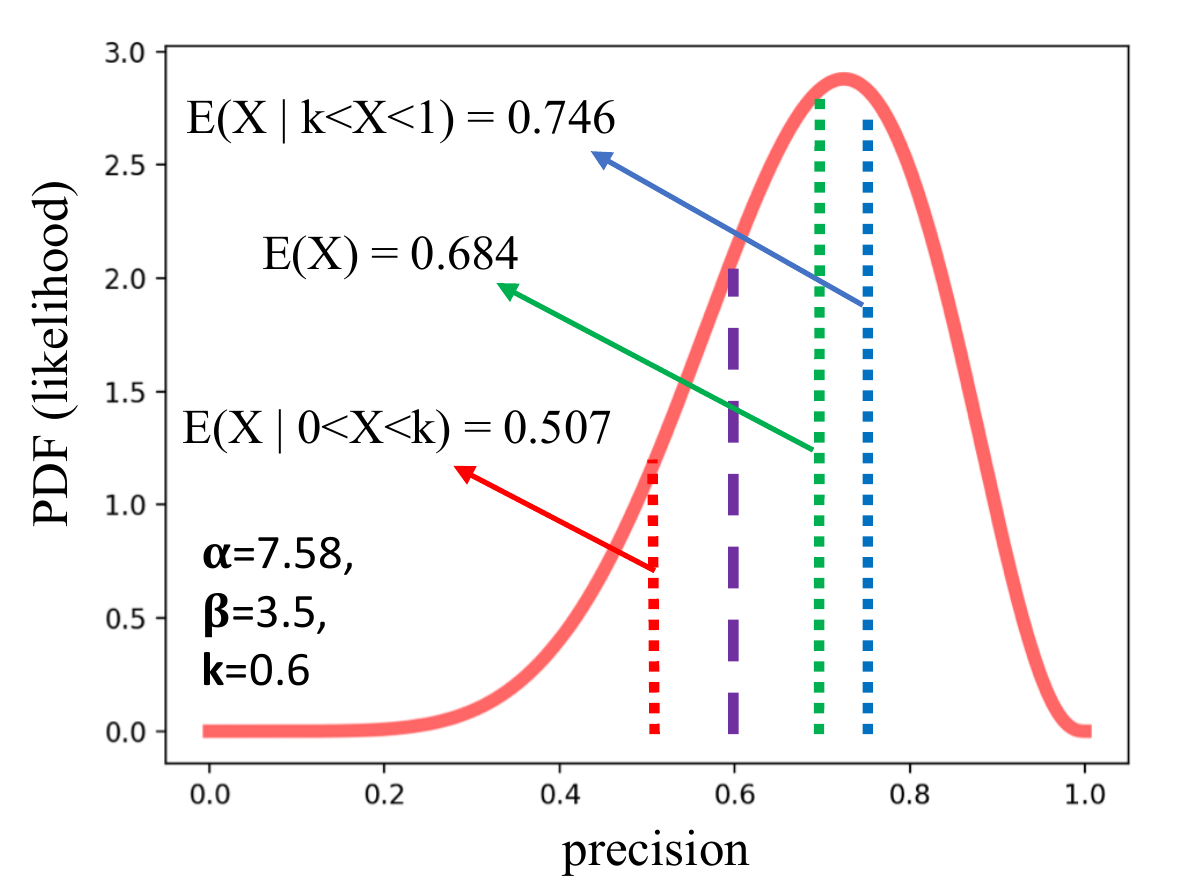} 
		\caption{Mean of truncated beta distribution.}
		\label{fig.upper lower mean}
	\end{center}		
\end{figure}
\begin{equation}
\label{eqn:lower mean}
E(X|0 < X < k) = \frac{\alpha}{\alpha+\beta}\frac{\int_{0}^{k} f(x;\alpha+1,\beta) dx}{\int_{0}^{k} f(x;\alpha,\beta) dx} 
\end{equation}
Following Equation~\eqref{eqn:lower mean}, we can calculate the upper mean of the truncated beta distribution to estimate the precision of the constraint-based bilingual lexicon induction when the agent transit to $s'_{sat}$ as
\begin{equation}
\label{eqn:upper mean}
E(X|k < X < 1) = \frac{\alpha}{\alpha+\beta}\frac{1-\int_{0}^{k} f(x;\alpha+1,\beta) dx}{1-\int_{0}^{k} f(x;\alpha,\beta) dx} 
\end{equation}

Using the same example in Section~\ref{sec:state transition}, using Equation~\eqref{eqn:mean}, Equation~\eqref{eqn:lower mean}, and Equation~\eqref{eqn:upper mean}, the beta distribution overall mean equals 0.684, lower mean equals 0.507, and upper mean equals 0.746 as shown in Figure~\ref{fig.upper lower mean}. Now we can estimate the size of the SATDict, $size(d_{(x,y):s})$ and the UnSATDict, $size(d_{(x,y):pu(z)})$ after taking pivot action with Equation~\eqref{eqn:SATDictSize} and Equation~\eqref{eqn:UnSATDictSize}, respectively.
\begin{equation}
\label{eqn:SATDictSize}
size(d_{(x,y):s}) = E(X|k < X < 1) \times size(d^c_{(x,y)})
\end{equation}
\begin{equation}
\label{eqn:UnSATDictSize}
size(d_{(x,y):pu(z)}) = E(X|0 < X < k) \times size(d^c_{(x,y)})
\end{equation}
\subsubsection{Value Iteration}
We use value iteration algorithm \cite{howard1960dynamic} to calculate utility (optimal policy) of each state by summing the cost for starting at state s and acting according to policies thereafter. Bellman~\cite{bellman2013dynamic}, via his Principle of Optimality, showed that the stochastic dynamic programming equation given below is guaranteed to find the optimal policy for the MDP. 
\begin{equation}
\label{eqn:value iteration}
V_{i}(s) = \begin{cases}
\min_{a \in A(s)} \sum_{s'} T(s,a,s')\big(C(s,a,s') + V_{i-1}(s')\big) & i>0 \\
0 & i=0
\end{cases}
\end{equation}
The above function, $V_i$, quantifies the long-term negative value, or cost, of reaching each state with $i$ actions remaining to be performed. Every state will have a policy of best action in order to minimize cumulative costs. Once we know the cost associated with each state of the plan, the optimal action for each state is the one which results in the minimum expected cost. In Equation~\eqref{eqn:policy} below, $\pi^*$ is the optimal policy which is simply a mapping from states to actions. Following the policy, we will obtain the optimal plan with the minimum cumulative costs.
\begin{equation}
\label{eqn:policy}
\pi^*(s) = \argmin_{a \in A(s)} \sum_{s'} T(s,a,s')\big(C(s,a,s') + V_{i-1}(s')\big) 
\end{equation}
\section{Experiment}
\begin{table}
	\small
	\caption{Similarity Matrix of The Target Languages}
	\label{tab:asjp}
	\begin{center}
		\begin{tabular}{lcccccc}
			\toprule
			Language&Indonesian&Javanese&Sundanese&Malay&Palembang Malay&Minangkabau\\
			\hline
			Javanese&24.09&& & & &\\
			Sundanese&39.43\%&21.82\%& & & &\\
			Malay&85.10\%&21.36\%&41.12\%& & &\\
			Palembang Malay&68.24\%&31.85\%&38.90\%&73.23\%&&\\
			Minangkabau&61.59\%&25.01\%&30.81\%&61.66\%&63.60\% &\\
			Banjarese Malay&71.57\%&32.5\%&38.72\%&70.93\%&63.53\%&60.39\%\\
			\bottomrule
		\end{tabular}
	\end{center}
\end{table}
To evaluate our MDP plan optimizer, we provide a sample experiment in Indonesia as part of Indonesia language sphere project~\cite{murakami2019indonesia}. To select target languages, we use an Automatic Similarity Judgment Program (ASJP) \cite{holman2011automated} following our previous work \cite{nasution2019simcluster}. Indonesia has 707 low-resource ethnic languages \cite{Lewis-15} that require our attention. There are two factors we consider in selecting the target languages: language similarity and number of speakers. In order to ensure that the induced bilingual dictionaries will be useful for many users, we listed the top 10 Indonesian ethnic languages ranked by the number of speakers. Since our constraint-based approach works better on closely related languages, we further generated the language similarity matrix by utilizing ASJP as shown in Table \ref{tab:asjp}. Based on number of speaker, we select Javanese and Sundanese. To find and coordinate native speakers of those languages, we collaborate with Telkom University. Based on relatedness with Indonesian, we select Malay, Minangkabau, Palembang Malay and Banjarese Malay. To find and coordinate native speakers of those language, we collaborate with Islamic University of Riau. Hence, we target 7 languages, i.e., Indonesian (ind), Malay (zlm), Minangkabau (min), Palembang Malay (plm), Banjarese Malay (bjn), Javanese (jav), and Sundanese (sun). We want to enrich/create the following dictionaries: $d_{(ind,zlm)}$, $d_{(ind,min)}$, $d_{(ind,bjn)}$, $d_{(ind,plm)}$, $d_{(ind,jav)}$, $d_{(ind,sun)}$, $d_{(zlm,min)}$, $d_{(zlm,bjn)}$, $d_{(zlm,plm)}$, $d_{(zlm,jav)}$, $d_{(zlm,sun)}$, $d_{(min,bjn)}$, $d_{(min,plm)}$, $d_{(min,jav)}$, $d_{(min,sun)}$, $d_{(bjn,plm)}$, $d_{(bjn,jav)}$, $d_{(bjn,sun)}$, $d_{(plm,jav)}$, $d_{(plm,sun)}$, and $d_{(jav,sun)}$ with at least 2,000 translation pairs each, $size(d^m_{(x,y)})=2,000$. To compare the effectiveness of the beta distribution model, we conducted two batch of experiments. The first batch of experiments includes 5 languages: Indonesian, Malay, Minangkabau, Javanese, and Sundanese with 10 combination of bilingual dictionaries: $d_{(ind,zlm)}$, $d_{(ind,min)}$, $d_{(ind,jav)}$, $d_{(ind,sun)}$, $d_{(zlm,min)}$, $d_{(zlm,jav)}$, $d_{(zlm,sun)}$, $d_{(min,jav)}$, $d_{(min,sun)}$, and $d_{(jav,sun)}$. The second batch of experiments includes two more languages which adds 11 combination of bilingual dictionaries: $d_{(ind,bjn)}$, $d_{(ind,plm)}$, $d_{(zlm,bjn)}$, $d_{(zlm,plm)}$, $d_{(min,bjn)}$, $d_{(min,plm)}$, $d_{(bjn,plm)}$, $d_{(bjn,jav)}$, $d_{(bjn,sun)}$, $d_{(plm,jav)}$, and $d_{(plm,sun)}$. In total, there are 21 combination of bilingual dictionaries created in this paper.

We model the $creationCost$ and $evaluationCost$ based on the availability of the native speakers. We provide example of modeling task for native speaker with Indonesian language families as target languages following our previous work \cite{nasution2018collab}. The detailed process of bilingual dictionaries generation process is explained in Algorithm~\ref{alg:bilingualDictionariesGeneration}.

\begin{algorithm}
	\SetAlgoNoLine
	\footnotesize
	\DontPrintSemicolon
	\KwIn{
		S, A, TS, T, C, dictionaryList \tcc*{output of Algorithm~\ref{alg:formalization}: State Transition Graph Generation}
	} 
	\KwOut{dictionaryList \tcc*{all combination of bilingual dictionaries from the targetLanguages}} 
	policy $\leftarrow$ valueIteration(S, A, TS, T, C);\tcc*{Calculating policy, a mapping from State to Action using Equation~\eqref{eqn:policy}}	
	state $\leftarrow$ S[0]; \tcc*{Start State}
	\While{state is not a finalState} 
	{
		action $\leftarrow$ policy.getAction(state);\;
		\If{action.getType() = investment}
		{					
			\tcc{CT1($L_{ind},L_x$): Creation and Evaluation of Indonesia-Ethnic Bilingual Dict}
			\If{$L_x$ or $L_y$ is Indonesian language $L_{ind}$}
			{					
				$d_{(x,y)}$ $\leftarrow$ invest($s_{(x,y)}$); \tcc*{create and evaluate the bilingual dictionary by a bilingual speaker}
				dictionaryList.updateSizeAndStatus($d_{(x,y)}$);					
			}					
			\tcc{CT2($L_x,L_y$): Creation and Evaluation of Ethnic-Ethnic Bilingual Dict}
			\Else
			{
				\If{native bilingual speaker $s_{(x,y)}$ is available}{
					$d_{(x,y)}$ $\leftarrow$ invest($s_{(x,y)}$); \tcc*{create and evaluate the bilingual dictionary by a bilingual speaker}
					dictionaryList.updateSizeAndStatus($d_{(x,y)}$);
				}
				\Else{
					$t_{(x,ind,y)}$ $\leftarrow$ invest($s_{(ind,x)}$, $s_{(ind,y)}$); \tcc*{create and evaluate the triple by two bilingual speakers}
					$d_{(x,y)}$ $\leftarrow$ induce($t_{(x,ind,y)}$);\;
					dictionaryList.updateSizeAndStatus($d_{(x,y)}$);
				}
			}
		}				
		\ElseIf{action.getType() = pivot}
		{
			$t_{(x,z,y)}$ $\leftarrow$ pivot($d_{(x,z)}$, $d_{(z,y)}$); \tcc*{use constraint-based bilingual lexicon induction}
			\tcc{T4($L_x,L_z,L_y$)}
			\If{native bilingual speaker $s_{(x,y)}$ is available}{
				$t_{(x,z,y)}$ $\leftarrow$ evaluate($t_{(x,z,y)}$, $s_{(x,y)}$); \tcc*{incorrect triples are pruned by a bilingual speaker}
				$d_{(x,y)}$$\leftarrow$ induce($t_{(x,z,y)}$);\;
				dictionaryList.updateSizeAndStatus($d_{(x,y)}$);
			}
			\Else{
				$t_{(x,z,y)}$ $\leftarrow$ evaluate($t_{(x,z,y)}$, $s_{(x,z)}$, $s_{(z,y)}$); \tcc*{incorrect triples are pruned by two bilingual speakers}
				induce $d_{(x,y)}$ from $t_{(x,z,y)}$;\;
				dictionaryList.updateSizeAndStatus($d_{(x,y)}$);
			}
		}			
		state $\leftarrow$ TS[state, action]; \tcc*{get the target state}
	}
	return dictionaryList;
	\caption{Bilingual Dictionaries Generation}
	\label{alg:bilingualDictionariesGeneration}
\end{algorithm}

\subsection{Modeling Task for Native Speaker}
Indonesian, a national language of Indonesia, is commonly used in both formal and informal settings, so, almost everyone can speak Indonesian well. However, to create bilingual dictionary $d_{(x,y)}$ between ethnic language $L_x$ and ethnic language $L_y$, there is a difficulty in finding a bilingual native speaker of the two ethnic languages. To overcome this limitation, we can firstly create triple $t_{(x,ind,y)}$ using the common language, Indonesian as pivot language $L_{ind}$ where $s_{(ind,x)}$, a native bilingual speaker of Indonesian language $L_{ind}$ - ethnic language $L_x$ and $s_{(ind,y)}$, a native bilingual speaker of Indonesian language $L_{ind}$ - ethnic language $L_y$ collaborate by explaining the senses with Indonesian language. Then, the bilingual dictionary $d_{(x,y)}$ can be induced from the triple $t_{(x,ind,y)}$. 

\begin{figure}[t!]
	\begin{center}
		\includegraphics[scale=0.48]{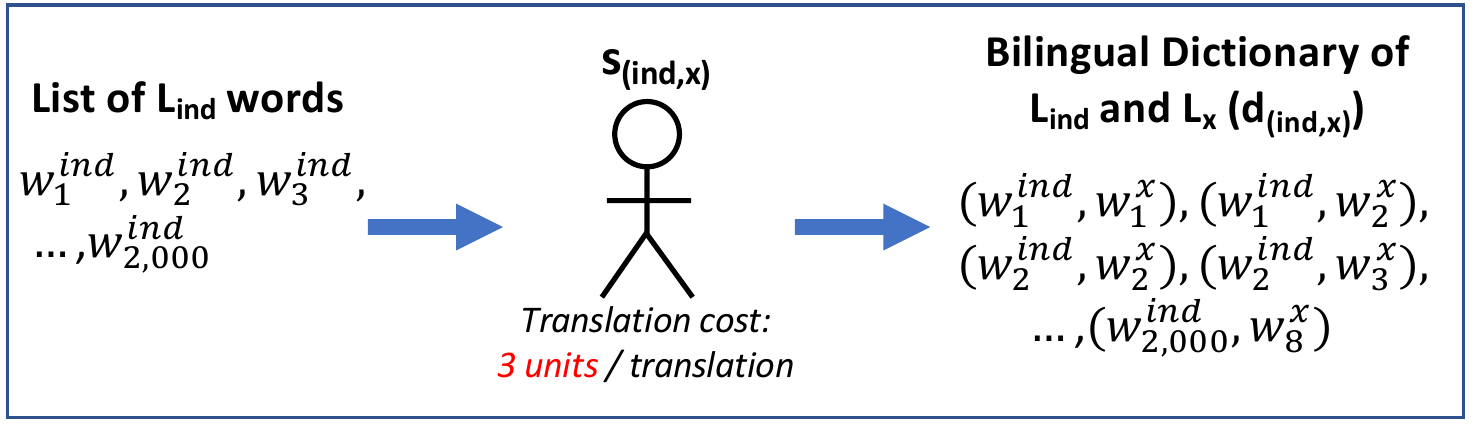}
		\caption{$T1(L_{ind},L_x)$: Creation of Bilingual Dictionary $d_{(ind,x)}$.}
		\label{fig:task1}
	\end{center}		
\end{figure}

\begin{figure}[t!]
	\begin{center}
		\includegraphics[scale=0.48]{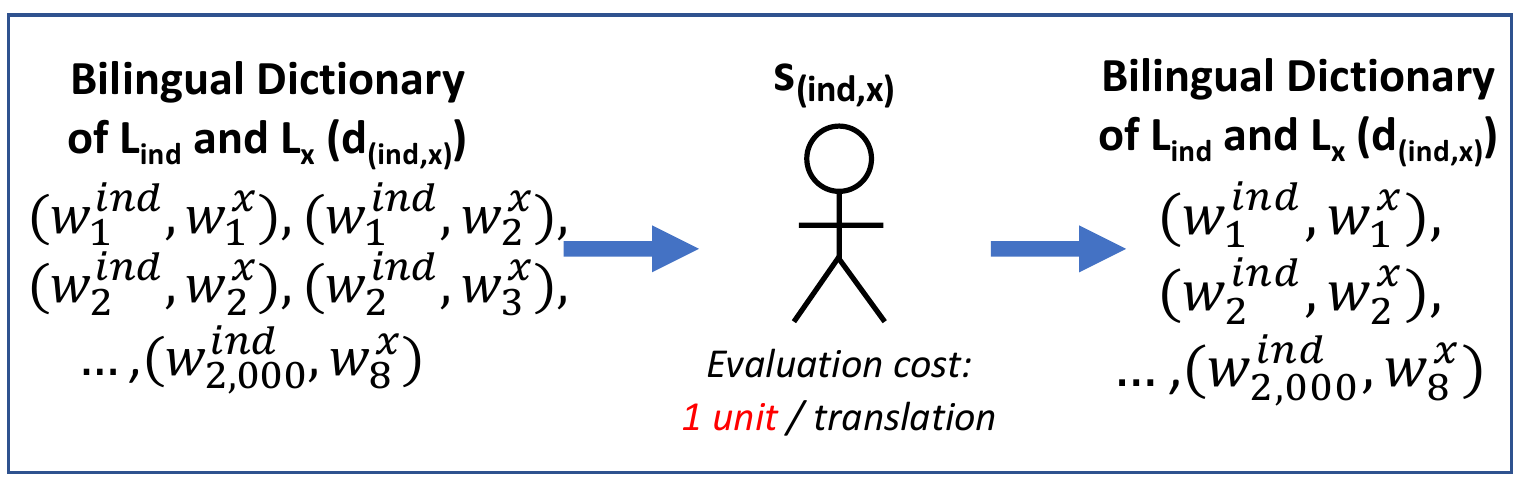}
		\caption{$T2(L_{ind},L_x)$: Evaluation of Bilingual Dictionary $d_{(ind,x)}$.}
		\label{fig:task2}
	\end{center}		
\end{figure}

We measure the cost of creation / evaluation for each translation with a unit time which is calculated from the estimated time taken for doing the task and average daily wages of student part-time worker in Indonesia. This unit time simply shows that the creation cost of bilingual dictionary $d_{(ind,x)}$ is three times it's evaluation cost as shown in Figure~\ref{fig:task1} and Figure~\ref{fig:task2}. When actually implementing our constraint-based bilingual lexicon induction, we need native speakers for manual creation of bilingual dictionaries or evaluation of the output dictionaries. We define several rules of which native speaker can create/evaluate which dictionary. A bilingual dictionary between ethnic language $L_{x}$ and ethnic language $L_{y}$, $d_{(x,y)}$ can be induced from a triple $t_{(x,ind,y)}$, while a triple $t_{(x,ind,y)}$ can be induced from a bilingual dictionary $d_{(ind,x)}$ and a bilingual dictionary $d_{(ind,y)}$. A bilingual dictionary between Indonesian language $L_{ind}$ and ethnic language $L_{x}$, $d_{(ind,x)}$ can be manually created or evaluated by a native bilingual speaker $s_{(ind,x)}$ as shown in Algorithm~\ref{alg:bilingualDictionariesGeneration} line number 6-9. A bilingual dictionary $d_{(x,y)}$ can be manually created or evaluated by a native bilingual speaker $s_{(ind,x)}$ and a native bilingual speaker $s_{(ind,y)}$ collaboratively as shown in Algorithm~\ref{alg:bilingualDictionariesGeneration} line number 15-19 or by a native bilingual speaker $s_{(x,y)}$ alone as shown in Algorithm~\ref{alg:bilingualDictionariesGeneration} line number 11-14. The incorrect triples $t_{(x,z,y)}$ output by the constraint-based bilingual lexicon induction are pruned by a native bilingual speaker $s_{(x,y)}$ individually as shown in Algorithm~\ref{alg:bilingualDictionariesGeneration} line number 24-28 or by a native bilingual speaker $s_{(x,z)}$ and a native bilingual speaker $s_{(z,y)}$ collaboratively as shown in Algorithm~\ref{alg:bilingualDictionariesGeneration} line number 29-33.
\begin{figure}[t!]
	\begin{center}
		\includegraphics[scale=0.45]{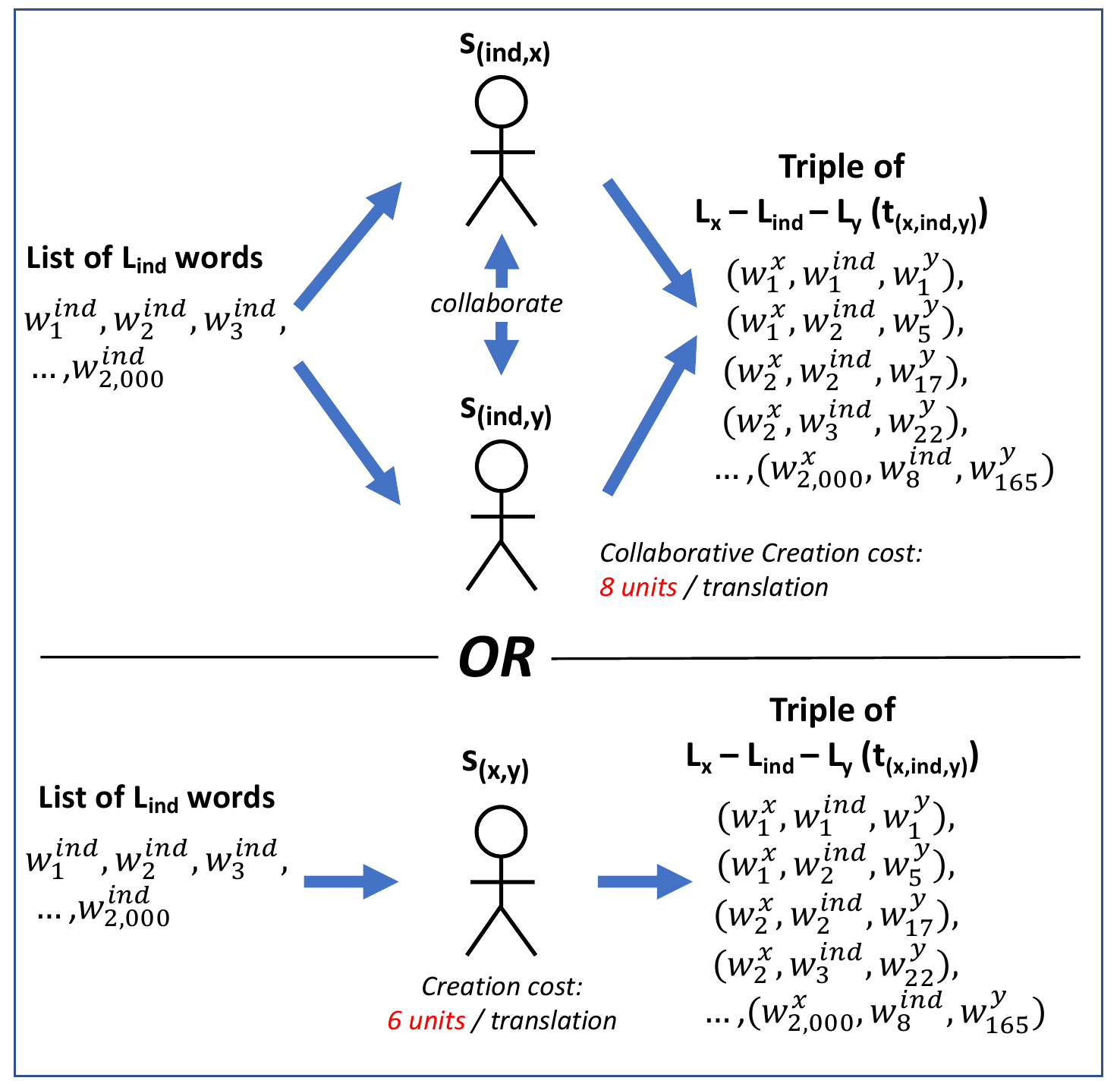}
		\caption{$T3(L_x,L_{ind},L_y)$: (Individual/Collaborative) Creation of Triple $t_{(x,ind,y)}$ to induce Bilingual Dictionary $d_{(x,y)}$.}
		\label{fig:task3}
	\end{center}		
\end{figure}

There are some bilingual dictionaries between Indonesian and Indonesian ethnic languages exist in a printed format. We may be able to digitalized the printed Indonesian - ethnic language bilingual dictionaries to a machine readable format. Nevertheless, when we connect the digitalized bilingual dictionary $d_{(ind,x)}$ and a bilingual dictionary $d_{(ind,y)}$ via Indonesian language $L_{ind}$ as a pivot, and further induced $d_{(x,y)}$ with our constraint-based approach, we expect that there will be many unreachable translation pair candidates since some Indonesian words in one bilingual dictionary may not exist in the other bilingual dictionary. In order to maximize the use of our pivot-based approach, we prepare a list of $2,000$ most commonly used Indonesian noun words to be translated to ethnic language $L_x$ to create a bilingual dictionary $d_{(ind,x)}$ by a native bilingual speaker $s_{(ind,x)}$ as shown in Figure \ref{fig:task1}. Due to budget limitation, we only allow the native speaker to translate an Indonesian word to up to five words of ethnic language $L_x$. 
\begin{figure}[t!]
	\begin{center}
		\includegraphics[scale=0.45]{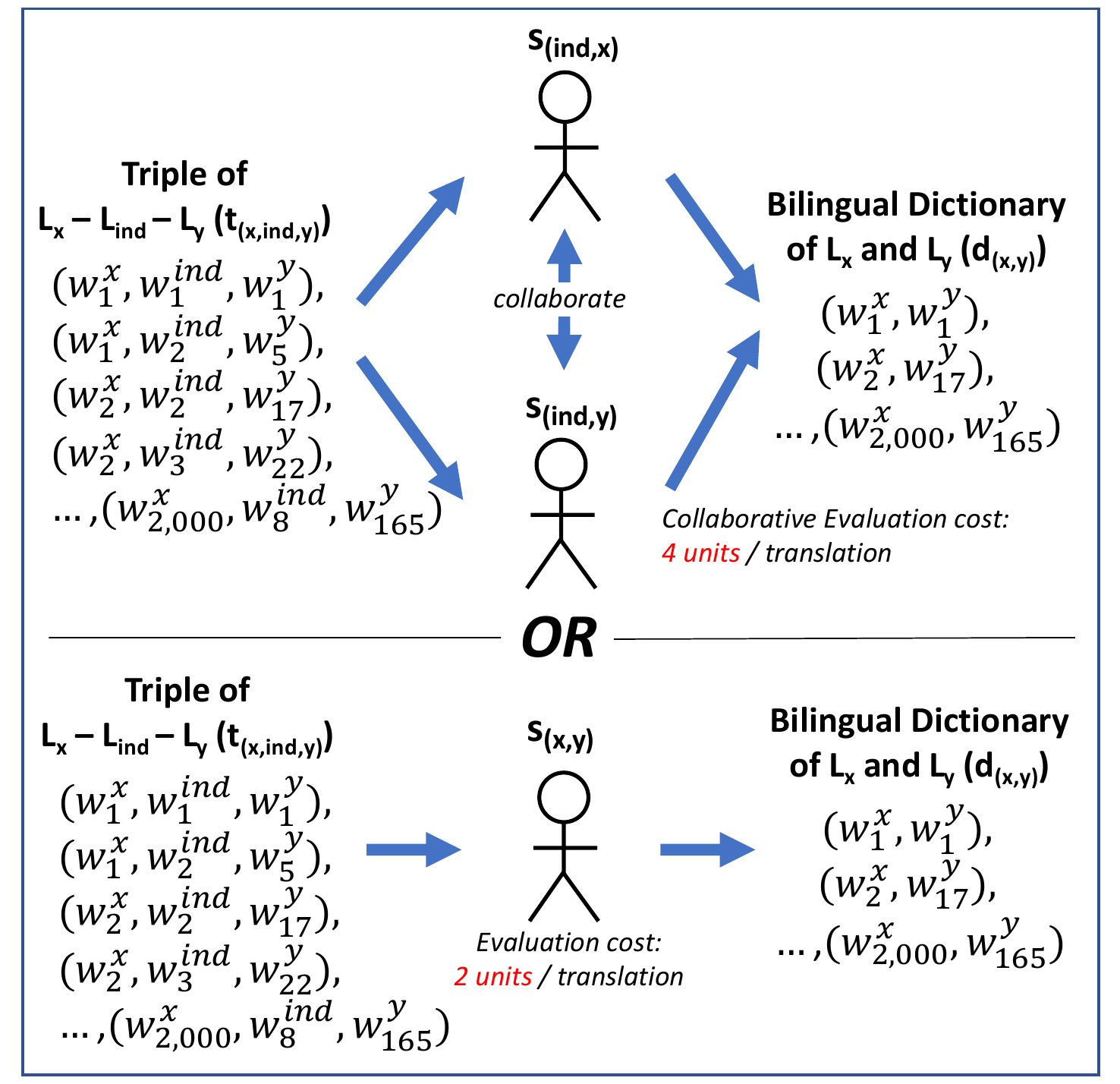}
		\caption{$T4(L_x,L_{ind},L_y)$: (Individual/Collaborative) Evaluation of Triple $t_{(x,ind,y)}$ to induce Bilingual Dictionary $d_{(x,y)}$.}
		\label{fig:task4}
	\end{center}		
\end{figure}

\begin{figure}[t!]
	\centering		
	\begin{subfigure}[b]{1\textwidth}
		\centering
		\includegraphics[scale=0.26]{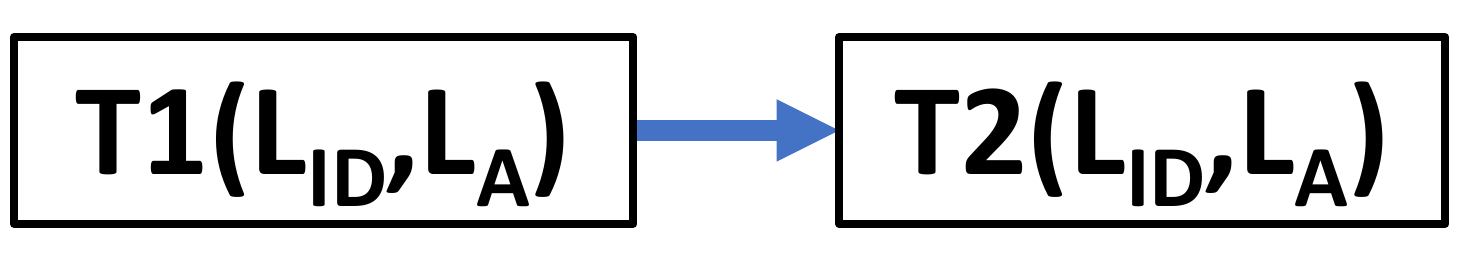}
		\caption{$CT1(L_{ind},L_x)$: Composite Task Creation and Evaluation of Bilingual Dictionary $d_{(ind,x)}$.}
		\label{fig:compositetask1}
	\end{subfigure}
	\begin{subfigure}[b]{1\textwidth}
		\centering
		\includegraphics[scale=0.26]{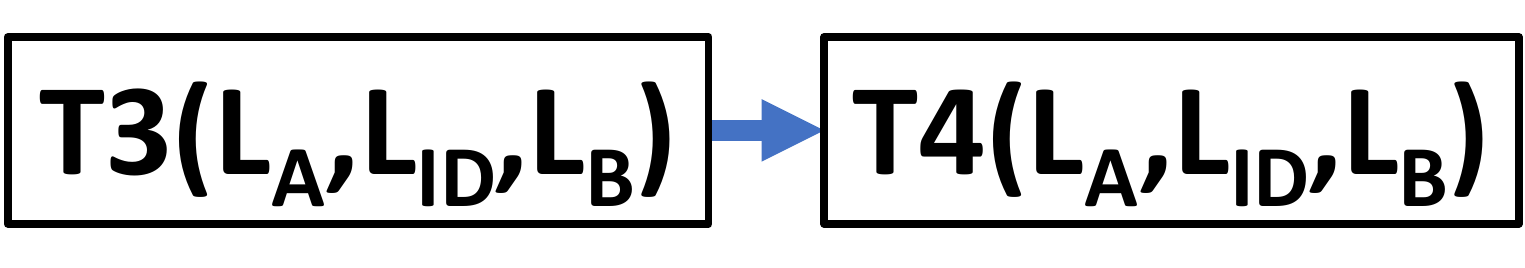}
		\caption{$CT2(L_x,L_{ind},L_y)$: Composite Task Creation and Evaluation of Bilingual Dictionary $d_{(x,y)}$.}
		\label{fig:compositetask2}
	\end{subfigure}
	\caption{Composite Tasks.}	
\end{figure}

To ensure the quality of the manually created bilingual dictionary $d_{(ind,x)}$, another native bilingual speaker $s_{(ind,x)}$ will evaluate the translation pairs as shown in Figure \ref{fig:task2}. We only pay correct translation pairs to the native bilingual speaker who do the creation task in order to motivate them to do the task carefully. To overcome the limitation in finding native bilingual speakers of two ethnic languages for creation and evaluation of bilingual dictionary $d_{(x,y)}$, two native bilingual speakers $s_{(ind,x)}$ and $s_{(ind,y)}$ can collaborate as shown in Figure \ref{fig:task3} and Figure \ref{fig:task4} respectively. Finally, there are two composite tasks, which are $CT1(L_{ind},L_x)$, a manual creation followed by evaluation of bilingual dictionary $d_{(ind,x)}$ as shown in Figure \ref{fig:compositetask1} and $CT2(L_x,L_{ind},L_y)$, a manual creation followed by evaluation of bilingual dictionary $d_{(x,y)}$ as shown in Figure \ref{fig:compositetask2}.

Finally, we integrate our constraint-based bilingual lexicon induction and plan optimizer with an online collaborative dictionary generation as a tool to bridge the spacial gap between native speakers \cite{nasution2018collab}.

\subsection{The First Batch of Experiments}
In the first batch of the experiment, $\alpha$ in beta-distribution represents language similarity between languages in the output dictionary such as language x and language y in $d_{(x,y)}$ as shown in Figure \ref{fig. dict_dependency}.
\begin{table}[t!]
	\caption{Estimated Cost of Actions following All Investment Plan}
	\label{tab:costHeuristicPlan}
	\small
	\centering
	\begin{threeparttable}
		\begin{center}
			\begin{tabular}{lrrr}
				\toprule
				Task Following Plan&\#Ordered Translation\tnote{1}&\#Paid Translation\tnote{2}&Total Cost (unit time)\\
				\hline
				CT1(ind, zlm) - 711 exist&1611&2900&5478\\
				CT1(ind, jav)&2500&4500&8500\\
				CT1(ind, sun)&2500&4500&8500\\
				CT2(zlm, min) - 1246 exist&943&1697&9802\\
				CT2(jav, sun)&2500&4500&26000\\
				CT2(zlm, jav)&2500&4500&26000\\
				CT2(min, sun)&2500&4500&26000\\
				CT2(zlm, sun)&2500&4500&26000\\
				CT2(min, jav)&2500&4500&26000\\
				\hline
				\textbf{TOTAL}&&&\textbf{162280}\\
				\bottomrule
			\end{tabular}
			\begin{tablenotes}
				\item[1] \textit{Estimating 0.8 human accuracy.}
				\item[2] \textit{\#Paid Translation = \#Created Translation + \#Evaluated Translation.}
			\end{tablenotes}
		\end{center}
	\end{threeparttable}	
\end{table}
\subsubsection{Plan Estimation}
\begin{figure}[t!]
	\centering		
	\begin{subfigure}[t]{0.32\textwidth}
		\centering
		\includegraphics[scale=0.4]{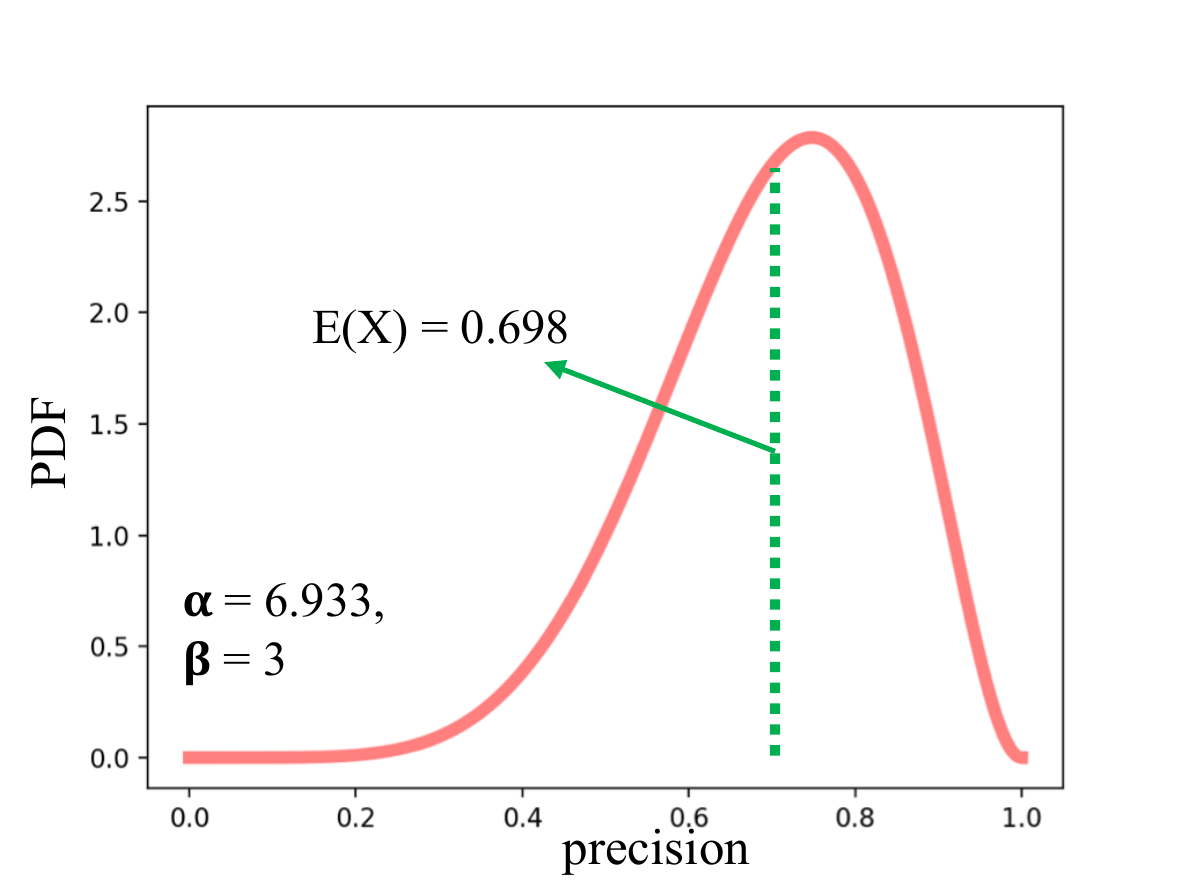}
		\caption{Malay-Minangkabau.}
		\label{fig:prior-zlm-min}
	\end{subfigure}
	\begin{subfigure}[t]{0.32\textwidth}
		\centering
		\includegraphics[scale=0.4]{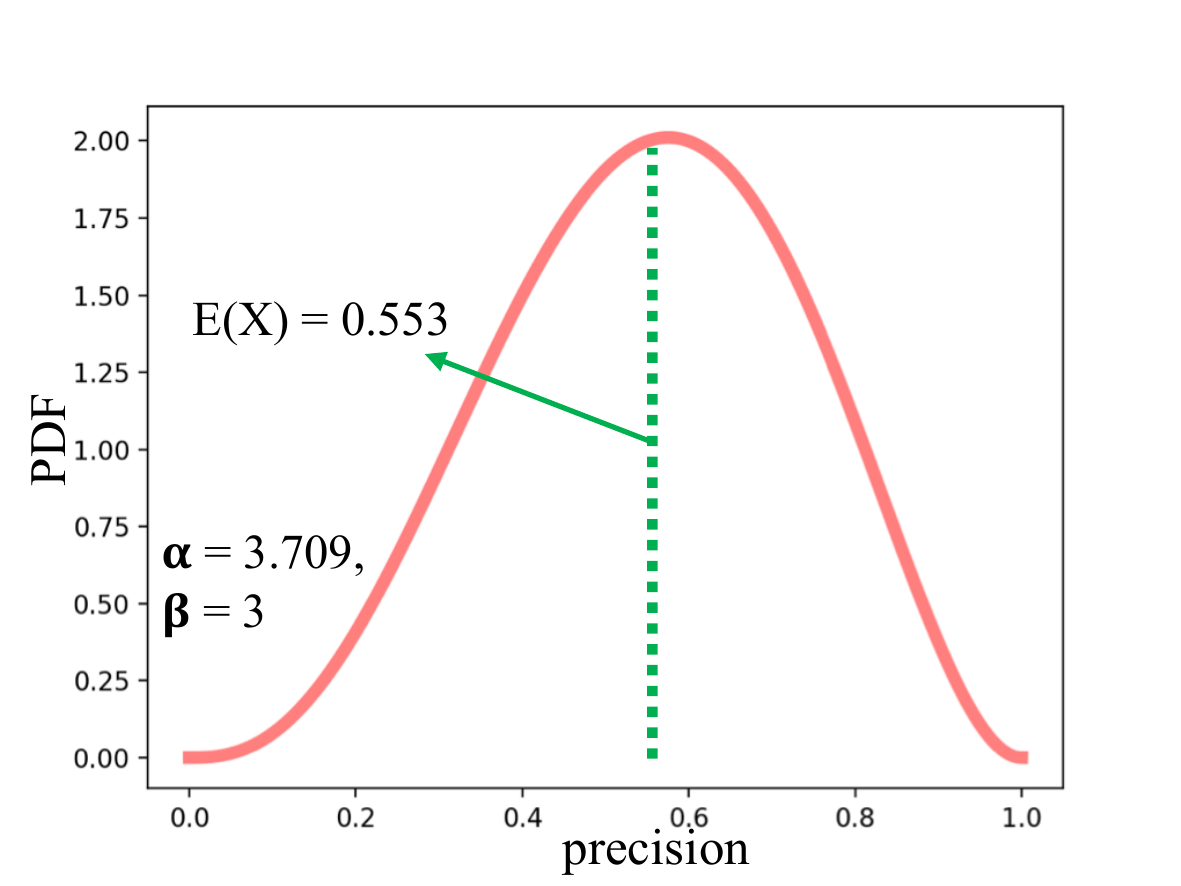}
		\caption{Malay-Javanese.}
		\label{fig:prior-zlm-jav}
	\end{subfigure}
	\begin{subfigure}[t]{0.32\textwidth}
		\centering
		\includegraphics[scale=0.4]{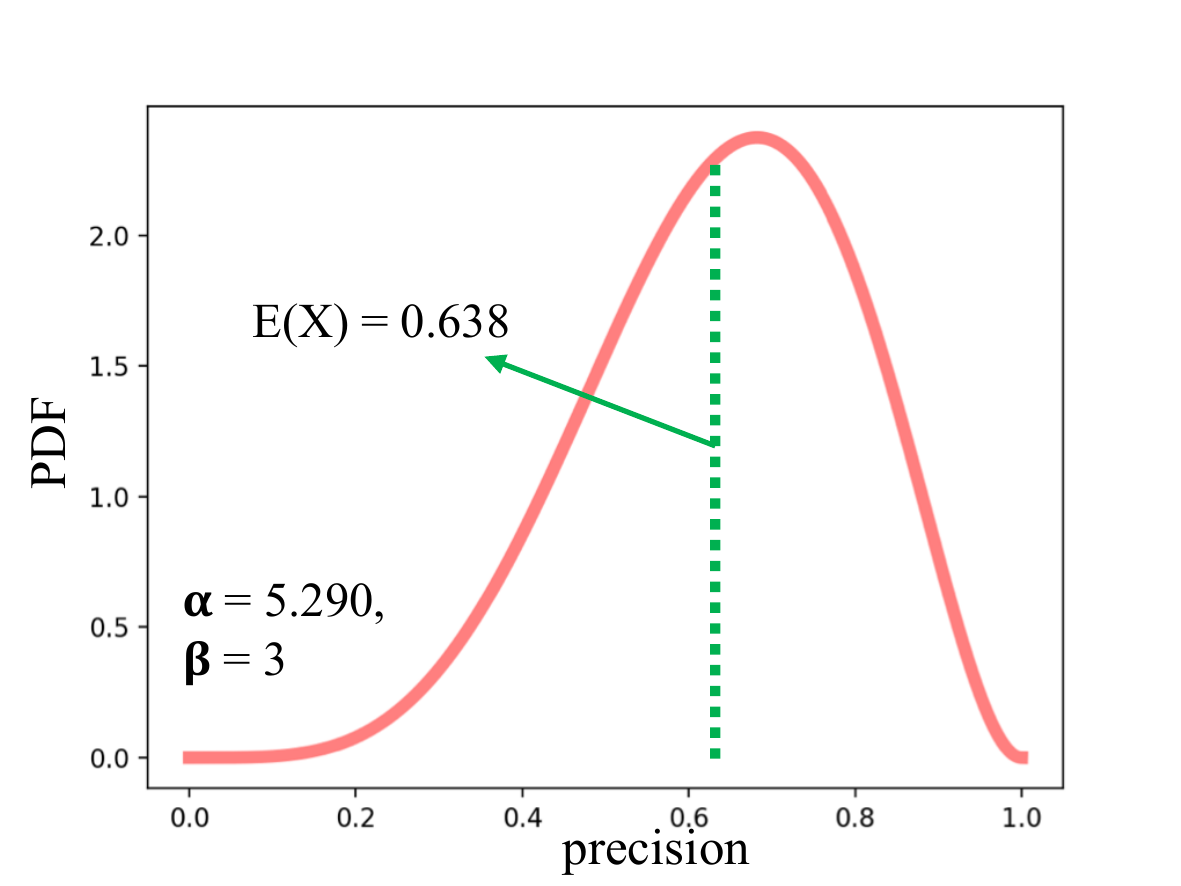}
		\caption{Malay-Sundanese.}
		\label{fig:prior-zlm-sun}
	\end{subfigure}
	\begin{subfigure}[t]{0.32\textwidth}
		\centering
		\includegraphics[scale=0.4]{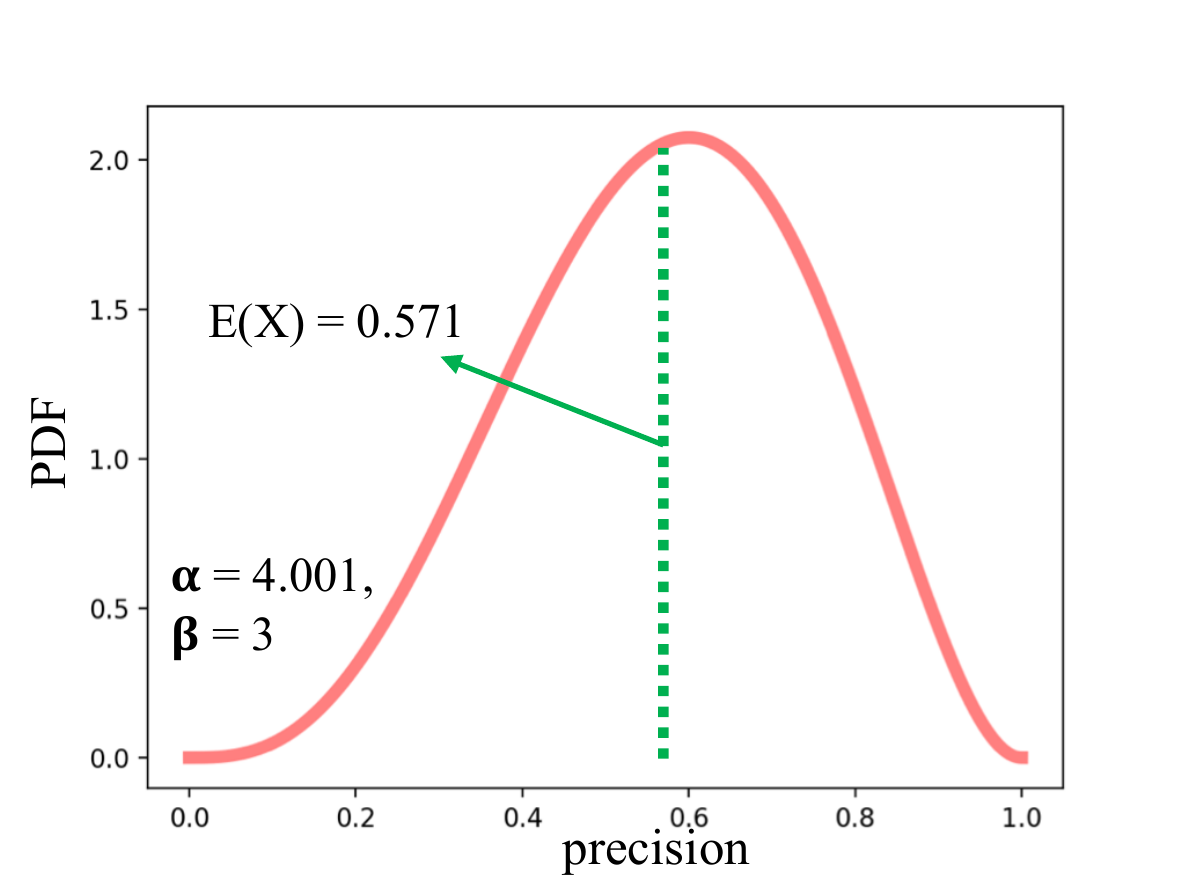}
		\caption{Minangkabau-Javanese.}
		\label{fig:prior-min-jav}
	\end{subfigure}
	\begin{subfigure}[t]{0.32\textwidth}
		\centering
		\includegraphics[scale=0.4]{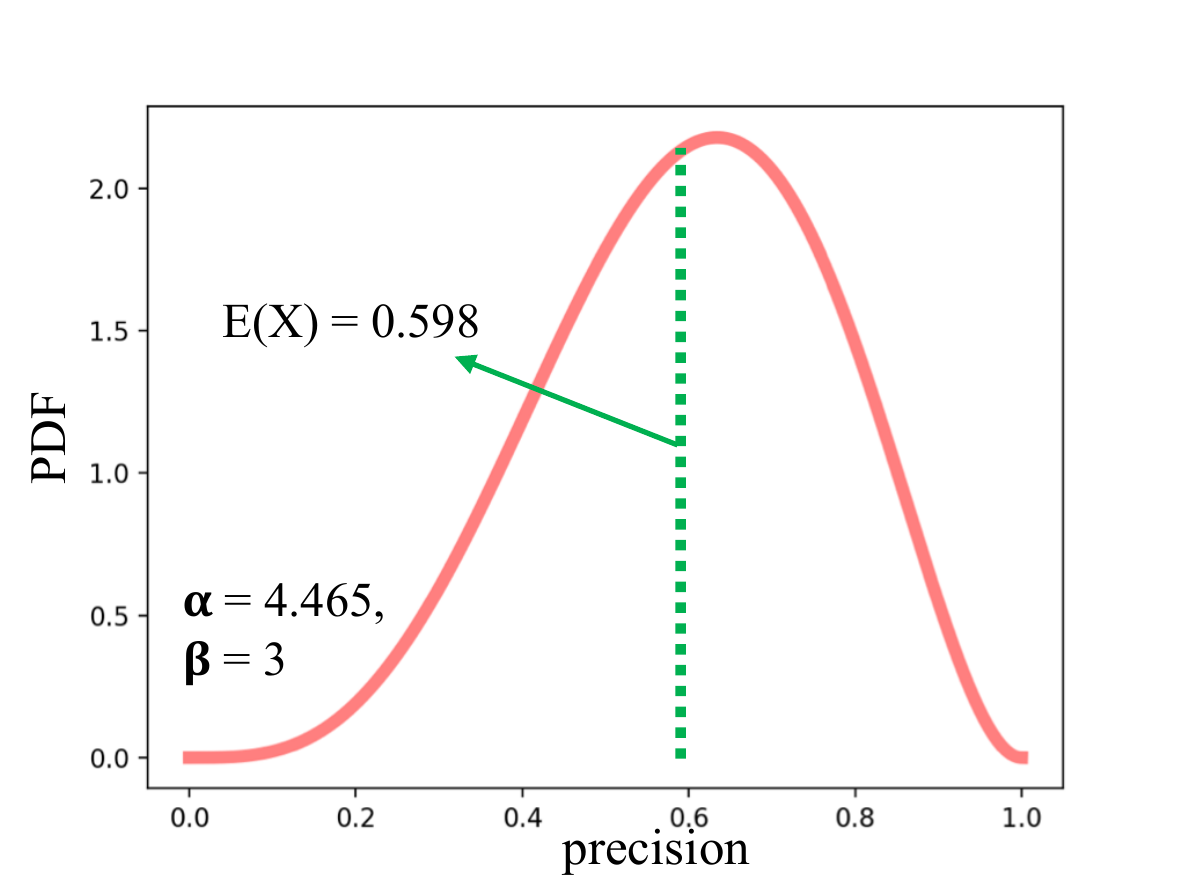}
		\caption{Minangkabau-Sundanese.}
		\label{fig:prior-min-sun}
	\end{subfigure}
	\begin{subfigure}[t]{0.32\textwidth}
		\centering
		\includegraphics[scale=0.4]{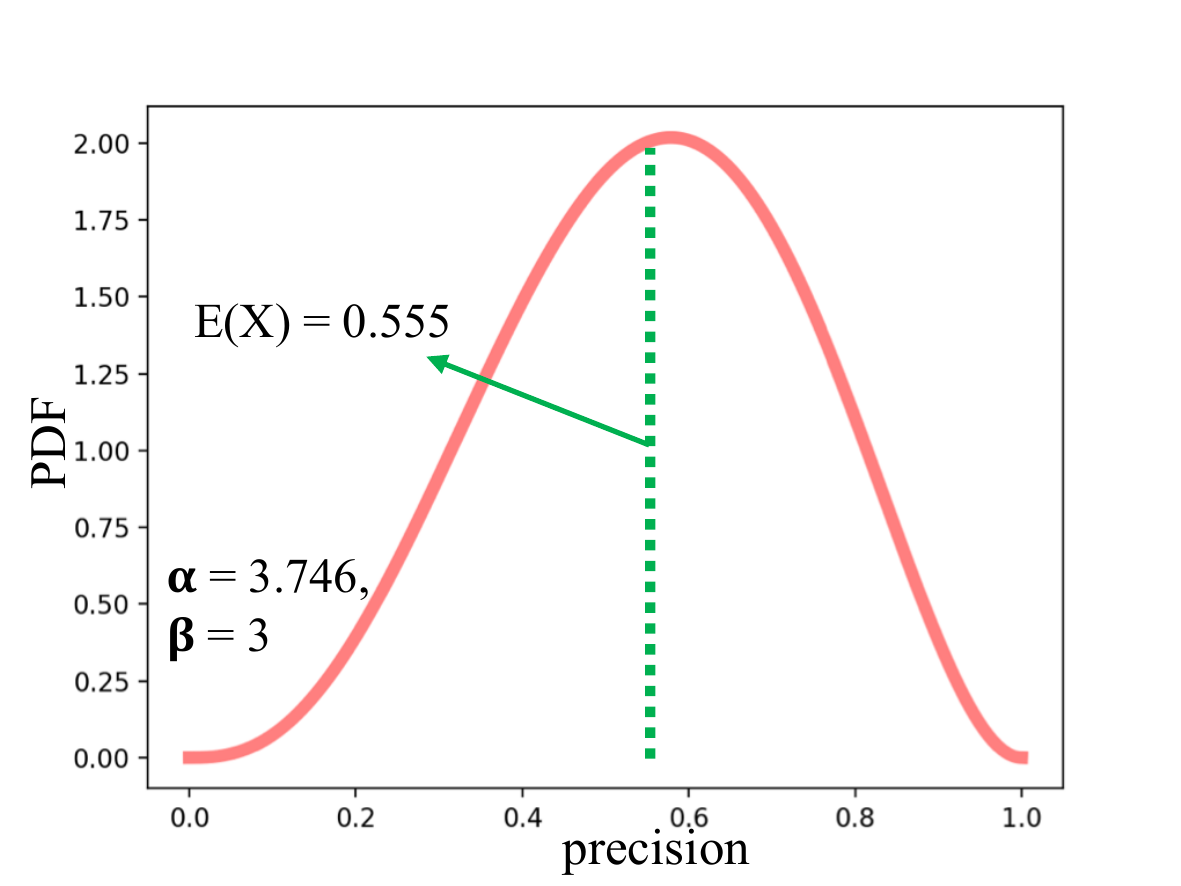}
		\caption{Javanese-Sundanese.}
		\label{fig:prior-jav-sun}
	\end{subfigure}
	\caption{Prior Beta Distribution for 6 Language Pairs.}	
\end{figure}

To show effectiveness of our method, we used, as a baseline, all investment plan as shown in Table~\ref{tab:costHeuristicPlan}. This all investment plan is just an estimation by simply calculating the number of translation pairs that need to be manually created and evaluated by human and then calculate each cost. We further constructed an estimated MDP optimal plan utilizing prior beta distributions of constraint-based bilingual lexicon induction precision for all language pairs that are generated by the constraint-based bilingual lexicon induction (aka, pivot action) as presented in Table~\ref{tab:costOptimalPlan}. We model $\alpha$ parameter from the language similarities shown in Figure~\ref{tab:asjp}. Since in practice, we predict that the topology in Figure~\ref{fig. polysemy}b is more likely to be generated, so, we model $\beta$ parameter by assuming all topology polysemy equals 3. We obtain the prior beta distributions as shown in Figure~\ref{fig:prior-zlm-min}-Figure~\ref{fig:prior-jav-sun} which are used to calculate the MDP state transition probability and cost function.

\begin{table}[t!]
	\caption{Estimated Cost of Actions following MDP Optimal Plan - The First Batch of Experiments}
	\label{tab:costOptimalPlan}
	\small
	\centering
	\begin{threeparttable}
		\begin{center}
			\begin{tabular}{lrrrrr}
				\toprule
				Task following Plan&\#Induced&Induction&Human&\#Paid&Total Cost\\
				&Translation&Precision\tnote{1}&Accuracy\tnote{2}&Translation\tnote{3}&(unit time)\\
				\hline
				CT1(ind, zlm) - 711 exist&&&0.8&2900&5478\\
				CT1(ind, jav)&&&0.8&4500&8500\\
				CT1(ind, sun)&&&0.8&4500&8500\\
				P(zlm, ind, min) - 1246 exist&2792&0.6981&&&0\\
				T4(zlm, ind, min)&&&1&2792&11170\\
				P(jav, ind, sun)&3285&0.6108&&&0\\
				T4(jav, ind, sun)&&&1&3285&13139\\
				P(zlm, ind, jav)&3283&0.6094&&&0\\
				T4(zlm, ind, jav)&&&1&3283&13134\\
				P(min, ind, sun)&2727&0.6817&&&0\\
				T4(min, ind, sun)&&&1&2727&10907\\
				P(zlm, ind, sun)&3644&0.6563&&&0\\
				T4(zlm, ind, sun)&&&1&3644&14578\\
				P(min, zlm, jav)&2694&0.6735&&&0\\
				T4(min, zlm, jav)&&&1&2694&10776\\
				\hline					
				\textbf{TOTAL}&&&&&\textbf{96182}\\
				\bottomrule
			\end{tabular}		
			\begin{tablenotes}
				\item[1] \textit{Estimated from beta distribution: language similarity as $\alpha$ and topology polysemy = 3 as $\beta$.}		
				\item[2] \textit{Human accuracy for creation task is estimated as 0.8 and 1 for evaluation task.}						
				\item[3] \textit{\#Paid Translation = \#Created Translation + \#Evaluated Translation.}
			\end{tablenotes}
		\end{center}
	\end{threeparttable}			
\end{table}

\subsubsection{Experiment Result}

\begin{table}[t!]
	\caption{Real Cost of Actions following MDP Optimal Plan - The First Batch of Experiments}
	\label{tab:realCostOptimalPlan}
	\footnotesize
	\centering
	\begin{threeparttable}
		\begin{center}
			\begin{tabular}{lrrrrrr}
				\toprule
				Task following Plan&Topology&\#Induced&Induction&Human&\#Paid&Total Cost\\
				&Polysemy\tnote{1}&Translation&Precision\tnote{2}&Accuracy\tnote{1}&Translation\tnote{3}&(unit time)\\
				\hline
				CT1(ind, zlm) - 711 exist&&&&0.868&3338&6440\\
				CT1(ind, jav)&&&&0.790&4573&8610\\
				CT1(ind, sun)&&&&0.830&4517&8615\\
				P(zlm, ind, min) - 1246 exist&3.355&1940&0.885&&&0\\
				T4(zlm, ind, min)&&&&1&1940&7760\\
				P(jav, ind, sun)&2.498&2071&0.824&&&0\\
				T4(jav, ind, sun)&&&&1&2071&8284\\
				CT2(jav, sun)&&&&0.838&715&4164\\
				P(zlm, ind, jav)&2.583&2018&0.801&&&0\\
				T4(zlm, ind, jav)&&&&1&2018&8072\\
				CT2(zlm, jav)&&&&0.843&892&5200\\
				P(min, ind, sun)&3.300&2239&0.802&&&0\\
				T4(min, ind, sun)&&&&1&2239&8956\\
				CT2(min, sun)&&&&0.732&435&2557\\
				P(zlm, ind, sun)&2.824&2029&0.833&&&0\\
				T4(zlm, ind, sun)&&&&1&2029&8116\\
				CT2(zlm, sun)&&&&0.840&665&3896\\
				P(min, zlm, jav)&3.192&2069&0.739&&&0\\
				T4(min, zlm, jav)&&&&1&2069&8276\\
				CT2(min, jav)&&&&0.957&678&4760\\
				\hline					
				\textbf{TOTAL}&&&&&&\textbf{93707}\tnote{4}\\
				\bottomrule
			\end{tabular}		
			\begin{tablenotes}
				\item[1] \textit{The average topology polysemy and human accuracy are close to our estimation in Table~\ref{tab:costOptimalPlan}.}		
				\item[2] \textit{All constraint-based bilingual lexicon induction precisions are higher than our estimation in Table~\ref{tab:costOptimalPlan}.}							
				\item[3] \textit{\#Paid Translation = \#Created Translation + \#Evaluated Translation.}
				\item[4] \textit{There are 42\% of cost reduction compared to the estimated all investment plan in Table~\ref*{tab:costHeuristicPlan} and 3\% of cost reduction compared to the estimated MDP optimal plan in Table~\ref{tab:costOptimalPlan}.}
			\end{tablenotes}
		\end{center}
	\end{threeparttable}			
\end{table}

The result depicted in Table~\ref{tab:realCostOptimalPlan} shows that our MDP optimal plan outperformed the all investment plan as regards of total cost with 42\% of cost reduction. The estimated total cost of actions following the MDP optimal plan shown in Table~\ref{tab:costOptimalPlan} is close to the total cost in the real experiment with 3\% of cost reduction. The average human accuracy shown in Table~\ref{tab:realCostOptimalPlan} is 0.837, close to our estimated human accuracy, 0.8. The average topology polysemy is 2.958, also close to our estimation, which is 3. 

\begin{figure}[t!]
	\begin{center}
		\includegraphics[scale=0.6]{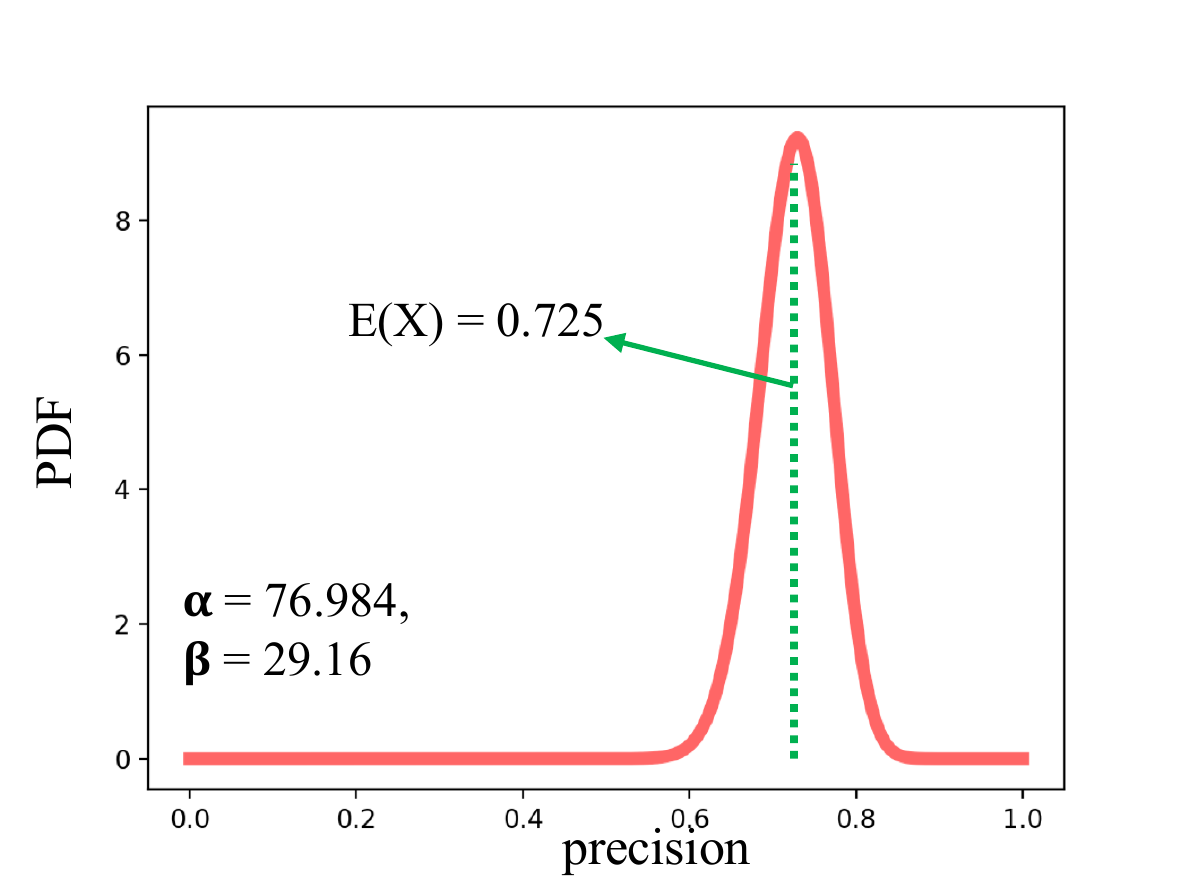}
		\caption{Final Posterior Beta Distribution of the First Batch of Experiments.}
		\label{fig:combined-6-posterior}
	\end{center}		
\end{figure}

From the experiment result, we can obtain the constraint-based bilingual lexicon induction precision. The likelihood's $\alpha$ parameter is calculated by normalizing the constraint-based bilingual lexicon induction precision to a range of [0, 10] and the $\beta$ parameter is $10-\alpha$.
	A posterior beta distribution can be constructed using Bayes' theorem as shown in Equation~\eqref{eqn:bayestheorem}. 
	\begin{equation}
	\label{eqn:bayestheorem}
	posterior \propto prior \times likelihood
	\end{equation}
	As shown in Table~\ref*{tab:priorposteriordistribution}, the posterior beta distribution $\alpha$ and $\beta$ parameters are calculated by adding the prior beta distribution $\alpha$ and $\beta$ parameters with the likelihood $\alpha$ and $\beta$ parameters. Since the likelihood's $\alpha$ and $\beta$ parameters are normalized to a range of [0, 10], close to the range of the prior beta distribution parameters [2, 10], the likelihood will contribute to adding believe toward the posterior beta distribution while not overwhelming the prior beta distribution. The final posterior beta distribution is obtained by multiplying all of the six posterior beta distributions shown in Table~\ref{tab:priorposteriordistribution} which can be used in the second batch of experiments. This final posterior beta distribution shown in Figure~\ref{fig:combined-6-posterior} represents the distribution of the constraint-based bilingual lexicon induction precision.

\begin{table}
	\caption{Prior and Posterior Beta Distribution of Pivot Action Precision - The First Batch of Experiments}
	\label{tab:priorposteriordistribution}
	\small
	\centering
	\begin{threeparttable}
		\begin{center}
			\begin{tabular}{lcccccccccc}
				\toprule
				\multirow{2}{*}{Language Pair}&Language&\multicolumn{3}{c}{Prior\tnote{1}}&\multicolumn{3}{c}{Likelihood\tnote{2}}&\multicolumn{3}{c}{Posterior\tnote{3}}\\
				&Similarity&$\alpha$&$\beta$&E(X)&$\alpha$&$\beta$&E(X)&$\alpha$&$\beta$&E(X)\\
				\hline
				zlm-min&0.617&6.933&3&0.698&8.85&1.15&0.885&15.783&4.15&0.792\\
				zlm-jav&0.214&3.709&3&0.553&8.01&1.99&0.801&11.719&4.99&0.701\\
				zlm-sun&0.411&5.290&3&0.638&8.33&1.67&0.833&13.62&4.67&0.745\\
				min-jav&0.250&4.001&3&0.571&7.39&2.61&0.739&11.391&5.61&0.670\\
				min-sun&0.308&4.465&3&0.598&8.02&1.98&0.802&12.485&4.98&0.715\\
				jav-sun&0.218&3.746&3&0.555&8.24&1.76&0.824&11.986&4.76&0.716\\
				\bottomrule
			\end{tabular}		
			\begin{tablenotes}
				\item[1] \textit{$\beta$ parameter is an initial believe because we predict that the topology in Figure~\ref{fig. polysemy}b is more likely to be generated, and $\alpha$ parameter is language similarity normalized to a range of [2, 10] to balance with the $\beta$ parameter.}
				\item[2] \textit{The likelihood's $\alpha$ parameter is calculated by normalizing the constraint-based bilingual lexicon induction precision to a range of [0, 10] and the $\beta$ parameter is $10-\alpha$.}						
				\item[3] \textit{The posterior beta distribution $\alpha$ and $\beta$ parameters are calculated by adding the prior beta distribution $\alpha$ and $\beta$ parameters with the likelihood $\alpha$ and $\beta$ parameters.}						
			\end{tablenotes}
		\end{center}
	\end{threeparttable}			
\end{table}


\subsection{The Second Batch of Experiments}
In the second batch of the experiment, $\alpha$ in beta-distribution represents average language similarity between input and output languages such as language x, language y, and language z in $d_{(x,z)}$, $d_{(z,y)}$, and $d_{(x,y)}$ as shown in Figure \ref{fig. dict_dependency}.

\begin{table}[t!]
	\caption{Estimated Cost of Actions following All Investment Plan - The Second Batch of Experiments}
	\label{tab:costHeuristicPlan2}
	\small
	\centering
	\begin{threeparttable}
		\begin{center}
			\begin{tabular}{lrrr}
				\toprule
				Task Following Plan&\#Ordered Translation\tnote{1}&\#Paid Translation\tnote{2}&Total Cost (unit time)\\
				\hline
				CT1(ind, bjn)&2500&4500&8500\\
				CT1(ind, plm)&2500&4500&8500\\
				CT2(bjn, zlm)&2500&4500&26000\\
				CT2(bjn, min)&2500&4500&26000\\
				CT2(bjn, jav)&2500&4500&26000\\
				CT2(bjn, sun)&2500&4500&26000\\
				CT2(bjn, plm)&2500&4500&26000\\
				CT2(plm, zlm)&2500&4500&26000\\
				CT2(plm, min)&2500&4500&26000\\
				CT2(plm, jav)&2500&4500&26000\\
				CT2(plm, sun)&2500&4500&26000\\
				\hline
				\textbf{TOTAL}&&&\textbf{251000}\\
				\bottomrule
			\end{tabular}
			\begin{tablenotes}
				\item[1] \textit{Estimating 0.8 human accuracy.}
				\item[2] \textit{\#Paid Translation = \#Created Translation + \#Evaluated Translation.}
			\end{tablenotes}
		\end{center}
	\end{threeparttable}	
\end{table}
\subsubsection{Plan Estimation}
We also used all investment plan as a baseline which is shown in Table~\ref{tab:costHeuristicPlan2}. We also estimated MDP optimal plan utilizing prior beta distributions the same way as presented in Table~\ref{tab:costOptimalPlan2}. We also model $\alpha$ parameter from the language similarities shown in Figure~\ref{tab:asjp} and model $\beta$ parameter by assuming all topology polysemy equals 3. However, we multiplied the beta distribution with the final posterior beta distribution of the First Batch of Experiments as shown in Figure~\ref{fig:combined-6-posterior}. We obtain the prior beta distributions as shown in Table~\ref{tab:priorposteriordistribution2} which are used to calculate the MDP state transition probability and cost function.

\begin{table}[t!]
	\caption{Estimated Cost of Actions following MDP Optimal Plan - The Second Batch of Experiments}
	\label{tab:costOptimalPlan2}
	\small
	\centering
	\begin{threeparttable}
		\begin{center}
			\begin{tabular}{lrrrrr}
				\toprule
				Task following Plan&\#Induced&Induction&Human&\#Paid&Total Cost\\
				&Translation&Precision\tnote{1}&Accuracy\tnote{2}&Translation\tnote{3}&(unit time)\\
				\hline
				CT1(ind, plm)&&&0.8&4500&8500\\
				CT1(ind, bjn)&&&0.8&4500&8500\\
				P(plm, ind, zlm)&1000&0.704&&&0\\
				T4(plm, ind, zlm)&&&1&1000&5000\\
				CT2(plm, zlm)&&&0.8&1000&7695.13\\
				P(bjn, ind, plm)&1000&0.669&&&0\\
				T4(bjn, ind, plm)&&&1&1000&5000\\
				CT2(bjn, plm)&&&0.8&1000&8595.6\\
				P(bjn, ind, min)&1000&0.645&&&0\\
				T4(bjn, ind, min)&&&1&1000&5000\\
				CT2(bjn, min)&&&0.8&1000&9225.67\\
				P(bjn, ind, zlm)&1500&0.758&&&\\
				T4(bjn, ind, zlm)&&&1&1500&7500\\
				CT2(bjn, zlm)&&&0.8&500&6274.67\\
				P(plm, bjn, min)&1000&0.625&&&\\
				T4(plm, bjn, min)&&&1&1000&5000\\
				CT2(plm, min)&&&0.8&1000&9750\\
				P(bjn, zlm, sun)&1000&0.503&&&\\
				T4(bjn, zlm, sun)&&&1&1000&5000\\
				CT2(bjn, sun)&&&0.8&1000&12933.26\\
				P(plm, ind, sun)&960&0.480&&&\\
				T4(plm, ind, sun)&&&1&960&4800\\
				CT2(plm, sun)&&&0.8&1040&13515.67\\
				P(bjn, ind, jav)&854&0.427&&&\\
				T4(bjn, ind, jav)&&&1&854&4270\\
				CT2(bjn, jav)&&&&1146&14892.8\\
				P(plm, bjn, jav)&852&0.426&&&\\
				T4(plm, bjn, jav)&&&1&852&4260\\
				CT2(plm, jav)&&&&1148&14917.07\\
				\hline					
				\textbf{TOTAL}&&&&&\textbf{160629.87}\\
				\bottomrule
			\end{tabular}		
			\begin{tablenotes}
				\item[1] \textit{Estimated from beta distribution (language similarity as $\alpha$ and topology polysemy = 3 as $\beta$) multiplied by the posterior beta distribution of the first batch of experiments.}		
				\item[2] \textit{Human accuracy for creation task is estimated as 0.8 and 1 for evaluation task.}						
				\item[3] \textit{\#Paid Translation = \#Created Translation + \#Evaluated Translation.}
			\end{tablenotes}
		\end{center}
	\end{threeparttable}			
\end{table}

\subsubsection{Experiment Result}
\begin{table}[t!]
	\caption{Real Cost of Actions following MDP Optimal Plan - The Second Batch of Experiments}
	\label{tab:realCostOptimalPlan2}
	\footnotesize
	\centering
	\begin{threeparttable}
		\begin{center}
			\begin{tabular}{lrrrrr}
				\toprule
				Task following Plan&\#Induced&Induction&Human&\#Paid&Total Cost\\
				&Translation&Precision\tnote{2}&Accuracy\tnote{1}&Translation\tnote{3}&(unit time)\\
				\hline
				CT1(ind, plm)&&&0.982&2079&8354\\
				CT1(ind, bjn)&&&0.986&2029&8144\\
				P(plm, ind, zlm)&1071&0.918&&&0\\
				T4(plm, ind, zlm)&&&1&1071&4284\\
				CT2(plm, zlm)&&&0.984&959&11572\\
				P(bjn, ind, plm)&1311&0.995&&&0\\
				T4(bjn, ind, plm)&&&1&1311&5244\\
				CT2(bjn, plm)&&&0.997&715&8588\\
				P(bjn, ind, min)&1165&0.858&&&0\\
				T4(bjn, ind, min)&&&1&1165&4660\\
				CT2(bjn, min)&&&0.969&853&10344\\
				P(bjn, ind, zlm)&1109&0.996&&&\\
				T4(bjn, ind, zlm)&&&1&1109&4436\\
				CT2(bjn, zlm)&&&0.992&897&10792\\
				P(plm, bjn, min)&946&0.893&&&\\
				T4(plm, bjn, min)&&&1&946&3784\\
				CT2(plm, min)&&&0.969&1069&12964\\
				P(bjn, zlm, sun)&1349&0.911&&&\\
				T4(bjn, zlm, sun)&&&1&1349&5396\\
				CT2(bjn, sun)&&&0.977&763&9228\\
				P(plm, ind, sun)&1178&0.969&&&\\
				T4(plm, ind, sun)&&&1&1178&4712\\
				CT2(plm, sun)&&&0.996&838&10068\\
				P(bjn, ind, jav)&1558&0.976&&&\\
				T4(bjn, ind, jav)&&&1&1558&6232\\
				CT2(bjn, jav)&&&0.81&447&5784\\
				P(plm, bjn, jav)&1055&0.967&&&\\
				T4(plm, bjn, jav)&&&1&1055&4220\\
				CT2(plm, jav)&&&0.932&1087&13360\\
				\hline					
				\textbf{TOTAL}&&&&&\textbf{152166}\tnote{4}\\
				\bottomrule
			\end{tabular}		
			\begin{tablenotes}
				\item[1] \textit{The average human accuracy is exceeding our estimation in Table~\ref{tab:costOptimalPlan2}.}		
				\item[2] \textit{All constraint-based bilingual lexicon induction precisions are higher than our estimation in Table~\ref{tab:costOptimalPlan2}.}							
				\item[3] \textit{\#Paid Translation = \#Created Translation + \#Evaluated Translation.}
				\item[4] \textit{There are 61.5\% of cost reduction compared to the estimated all investment plan in Table~\ref*{tab:costHeuristicPlan2} and 39.4\% of cost reduction compared to the estimated MDP optimal plan in Table~\ref{tab:costOptimalPlan2}.}
			\end{tablenotes}
		\end{center}
	\end{threeparttable}			
\end{table}
The result depicted in Table~\ref{tab:realCostOptimalPlan2} shows that our MDP optimal plan outperformed the all investment plan as regards of total cost with 61.5\% of cost reduction. The estimated total cost of actions following the MDP optimal plan shown in Table~\ref{tab:costOptimalPlan2} is close to the total cost in the real experiment with 39.4\% of cost reduction. The average human accuracy shown in Table~\ref{tab:realCostOptimalPlan2} is 0.963, exceeding our estimated human accuracy, 0.8.

\begin{figure}[t!]
	\begin{center}
		\includegraphics[scale=0.6]{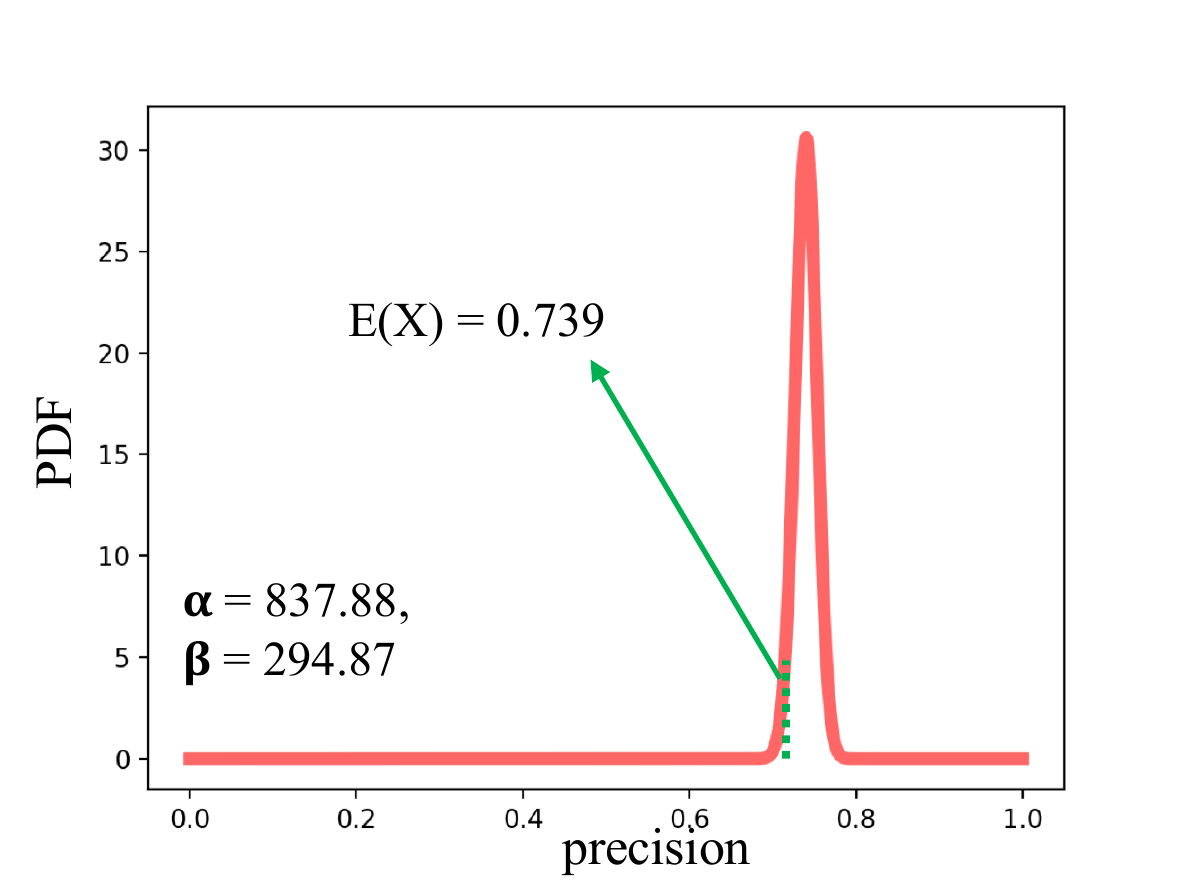}
		\caption{Final Posterior Beta Distribution for the Second Batch of Experiments.}
		\label{fig:combined-9-posterior}
	\end{center}		
\end{figure}

From the experiment result, the likelihood's $\alpha$ parameter and the $\beta$ parameter are obtained, then the posterior beta distribution are also constructed. As shown in Table~\ref*{tab:priorposteriordistribution2}, the posterior beta distribution $\alpha$ and $\beta$ parameters are calculated by adding the prior beta distribution $\alpha$ and $\beta$ parameters with the likelihood $\alpha$ and $\beta$ parameters. The final posterior beta distribution is obtained by multiplying all of the six posterior beta distribution shown in Table~\ref{tab:priorposteriordistribution2} which can be used in the future experiments. This final posterior beta distribution shown in Figure~\ref{fig:combined-9-posterior} represents the latest distribution of the constraint-based bilingual lexicon induction precision.

\begin{table}
	\caption{Prior and Posterior Beta Distribution of Pivot Action Precision - The Second Batch of Experiments}
	\label{tab:priorposteriordistribution2}
	\small
	\centering
	\begin{threeparttable}
		\begin{center}
			\begin{tabular}{lcccccccccc}
				\toprule
				\multirow{2}{*}{Language Triple}&Avg Language&\multicolumn{3}{c}{Prior\tnote{1}}&\multicolumn{3}{c}{Likelihood\tnote{2}}&\multicolumn{3}{c}{Posterior\tnote{3}}\\
				&Similarity&$\alpha$&$\beta$&E(X)&$\alpha$&$\beta$&E(X)&$\alpha$&$\beta$&E(X)\\
				\hline
				plm-ind-zlm&0.755&85.026&32.160&0.725&9.760&0.240&0.976&94.786&32.400&0.745\\
				bjn-ind-plm&0.678&84.406&32.160&0.724&9.960&0.040&0.996&94.366&32.200&0.746\\
				bjn-ind-min&0.645&84.145&32.160&0.723&9.690&0.310&0.969&93.835&32.470&0.743\\
				bjn-ind-zlm&0.759&85.053&32.160&0.726&9.180&0.820&0.918&94.233&32.980&0.741\\
				plm-bjn-min&0.625&83.985&32.160&0.723&9.110&0.890&0.911&93.095&33.050&0.738\\
				bjn-zlm-sun&0.503&83.005&32.160&0.721&9.690&0.310&0.969&92.695&32.470&0.741\\
				plm-ind-sun&0.489&82.893&32.160&0.720&8.580&1.420&0.858&91.473&33.580&0.731\\
				bjn-ind-jav&0.427&82.402&32.160&0.719&8.930&1.070&0.893&91.332&33.230&0.733\\
				plm-bjn-jav&0.426&82.394&32.160&0.719&9.670&0.330&0.967&92.064&32.490&0.739\\
				\bottomrule
			\end{tabular}		
			\begin{tablenotes}
				\item[1] \textit{$\beta$ parameter is an initial believe because we predict that the topology in Figure~\ref{fig. polysemy}b is more likely to be generated, and $\alpha$ parameter is language similarity normalized to a range of [2, 10] to balance with the $\beta$ parameter.}
				\item[2] \textit{The likelihood's $\alpha$ parameter is calculated by normalizing the constraint-based bilingual lexicon induction precision to a range of [0, 10] and the $\beta$ parameter is $10-\alpha$.}						
				\item[3] \textit{The posterior beta distribution $\alpha$ and $\beta$ parameters are calculated by adding the prior beta distribution $\alpha$ and $\beta$ parameters with the likelihood $\alpha$ and $\beta$ parameters.}						
			\end{tablenotes}
		\end{center}
	\end{threeparttable}			
\end{table}

\section{Discussion}
The result of the second batch of experiments outperformed the result of the first batch of experiments. In the first batch of experiments, there are 42\% of cost reduction compared to the estimated all investment plan and 3\% of cost reduction compared to the estimated MDP optimal plan, while in the second batch of experiments, there are 61.5\% of cost reduction compared to the estimated all investment plan and 39.4\% of cost reduction compared to the estimated MDP optimal plan. This shows that the experimental design in the second batch of experiments is potential to be used in the future works. The $\alpha$ in beta-distribution should represents average language similarity between input and output languages such as language x, language y, and language z in $d_{(x,z)}$, $d_{(z,y)}$, and $d_{(x,y)}$. Utilizing the final posterior beta distribution of the first batch of experiments to construct prior beta distribution of the second batch of experiments has been proven to be useful to help the MDP to estimate the optimal plan.

The current plan optimization algorithm is static/offline as the policy is only calculated once in Algorithm~\ref{alg:bilingualDictionariesGeneration} line number 1. After executing one or two actions from the static optimal plan, the previously optimal plan can be sub-optimal. For example, in our estimated MDP optimal plan shown in Table~\ref{tab:costOptimalPlan}, all pivot action successfully induced bilingual dictionaries with a satisfying size, however, after following the MDP optimal plan, despite of the higher constraint-based bilingual lexicon induction precision compared to the estimation, only one out of six pivot actions successfully induced bilingual dictionaries with a satisfying size. This phenomena is due to the error in estimating the size of translation pair candidates. We estimated that all average polysemy of the topology will be medium as shown in Figure~\ref{fig. polysemy}(b) while in reality, we can find a lot of transgraph with a one-to-one relation with the lowest average polysemy of the topology as shown in Figure~\ref{fig. polysemy}(a).

To make a dynamic/online plan optimization, we can update Algorithm~\ref{alg:bilingualDictionariesGeneration} by adding a recursive procedure to re-formalize the problem with Algorithm~\ref{alg:formalization} with updated information of the environment (size of translation pair candidates and dictionary status) every time after executing an action based on the current policy and further re-execute the new policy. This will make the planOptimizer adaptable to the changing of the environment. With a dynamic plan optimization, we can get a better estimation as well as reducing the computational complexity of the problem since the variable and the corresponding domain will be greatly reduced as more action has been executed, in other word, the number of states and actions generated by Algorithm~\ref{alg:formalization} will be greatly reduced. 

There is also a possibility to relax One-Time Induction Constraint ($C_3$) into a soft-constraint. However, this could lead to an overlapped result when more than one constraint-based bilingual lexicon induction taken with different pivot languages. A discount parameter can be introduce to estimate the degree of overlapping result.
\section{Conclusion}
Despite the great potential of our constraint-based bilingual lexicon induction to enrich low-resource languages with machine readable bilingual dictionaries as the sole input, when one wants to acquire every possible combination of bilingual dictionaries from the language set with a minimum dictionary size predefined but some input dictionaries are small, it is difficult to construct an optimal plan in which the order of executing dictionary creation methods including the manual creation by human will yield the least total cost to be paid. Our MDP model can calculate the cumulative cost while predicting and considering the probability of the constraint-based method yielding a satisfying output bilingual dictionary as utility for every state to get a better prediction of the most feasible optimal plan. \par
Our key research contribution is a twofold. For the earliest implementation of our approach, a prior beta distribution of constraint-based bilingual lexicon induction precision is modeled with language similarity and topology polysemy as $\alpha$ and $\beta$ parameters, respectively. After one episode of experiment, a posterior beta distribution can be constructed by utilizing the constraint-based bilingual lexicon induction precision as an added believe to the prior beta distribution while not overwhelming the prior beta distribution. The second key research contribution is the MDP optimal plan formalization itself. Our formalization allows user to get estimation of the feasible optimal plan with the least total cost before actually implementing the constraint-based bilingual lexicon induction in a big scale. The final posterior beta distribution of the second batch of experiments should be utilized to construct prior beta distribution for the future experiments.\par
In our future work, we will discuss about the dynamic/online plan optimization. There is also a possibility to relax One-Time Induction Constraint ($C_3$) into a soft-constraint. However, this could lead to an overlapped result when more than one constraint-based bilingual lexicon induction taken with different pivot languages. A discount parameter can be introduce to estimate the degree of overlapping result.

\begin{acks}	
	This research was partially supported by a Grant-in-Aid for Scientific Research (A) (17H00759, 2017-2020) and a Grant-in-Aid for Young Scientists (A) (17H04706, 2017-2020) from Japan Society for the Promotion of Science (JSPS). This research was partially supported by Universitas Islam Riau (UIR) and Universiti Teknologi PETRONAS (UTP) Joint Research Program. The first author was supported by Indonesia Endownment Fund for Education (LPDP).
\end{acks}

\bibliographystyle{ACM-Reference-Format}
\bibliography{sample-base}


\begin{thebibliography}{30}


\ifx \showCODEN    \undefined \def \showCODEN     #1{\unskip}     \fi
\ifx \showDOI      \undefined \def \showDOI       #1{#1}\fi
\ifx \showISBNx    \undefined \def \showISBNx     #1{\unskip}     \fi
\ifx \showISBNxiii \undefined \def \showISBNxiii  #1{\unskip}     \fi
\ifx \showISSN     \undefined \def \showISSN      #1{\unskip}     \fi
\ifx \showLCCN     \undefined \def \showLCCN      #1{\unskip}     \fi
\ifx \shownote     \undefined \def \shownote      #1{#1}          \fi
\ifx \showarticletitle \undefined \def \showarticletitle #1{#1}   \fi
\ifx \showURL      \undefined \def \showURL       {\relax}        \fi
\providecommand\bibfield[2]{#2}
\providecommand\bibinfo[2]{#2}
\providecommand\natexlab[1]{#1}
\providecommand\showeprint[2][]{arXiv:#2}

\bibitem[\protect\citeauthoryear{Ans{\'o}tegui, Bonet, and Levy}{Ans{\'o}tegui
  et~al\mbox{.}}{2009}]%
        {ansotegui2009solving}
\bibfield{author}{\bibinfo{person}{Carlos Ans{\'o}tegui},
  \bibinfo{person}{Mar{\'\i}a~Luisa Bonet}, {and} \bibinfo{person}{Jordi
  Levy}.} \bibinfo{year}{2009}\natexlab{}.
\newblock \showarticletitle{Solving (weighted) partial MaxSAT through
  satisfiability testing}.
\newblock In \bibinfo{booktitle}{\emph{Theory and Applications of
  Satisfiability Testing-SAT 2009}}. \bibinfo{publisher}{Springer},
  \bibinfo{pages}{427--440}.
\newblock


\bibitem[\protect\citeauthoryear{Bellman}{Bellman}{2013}]%
        {bellman2013dynamic}
\bibfield{author}{\bibinfo{person}{Richard Bellman}.}
  \bibinfo{year}{2013}\natexlab{}.
\newblock \bibinfo{booktitle}{\emph{Dynamic programming}}.
\newblock \bibinfo{publisher}{Courier Corporation}.
\newblock


\bibitem[\protect\citeauthoryear{Brown, Cocke, Pietra, Pietra, Jelinek,
  Lafferty, Mercer, and Roossin}{Brown et~al\mbox{.}}{1990}]%
        {brown1990statistical}
\bibfield{author}{\bibinfo{person}{Peter~F Brown}, \bibinfo{person}{John
  Cocke}, \bibinfo{person}{Stephen A~Della Pietra}, \bibinfo{person}{Vincent
  J~Della Pietra}, \bibinfo{person}{Fredrick Jelinek}, \bibinfo{person}{John~D
  Lafferty}, \bibinfo{person}{Robert~L Mercer}, {and} \bibinfo{person}{Paul~S
  Roossin}.} \bibinfo{year}{1990}\natexlab{}.
\newblock \showarticletitle{A statistical approach to machine translation}.
\newblock \bibinfo{journal}{\emph{Computational linguistics}}
  \bibinfo{volume}{16}, \bibinfo{number}{2} (\bibinfo{year}{1990}),
  \bibinfo{pages}{79--85}.
\newblock


\bibitem[\protect\citeauthoryear{Doshi, Goodwin, Akkiraju, and Verma}{Doshi
  et~al\mbox{.}}{2004}]%
        {doshi2004workflowcomposition}
\bibfield{author}{\bibinfo{person}{P. Doshi}, \bibinfo{person}{R. Goodwin},
  \bibinfo{person}{R. Akkiraju}, {and} \bibinfo{person}{K. Verma}.}
  \bibinfo{year}{2004}\natexlab{}.
\newblock \showarticletitle{Dynamic workflow composition using Markov decision
  processes}. In \bibinfo{booktitle}{\emph{Proceedings. IEEE International
  Conference on Web Services, 2004.}} \bibinfo{pages}{576--582}.
\newblock
\urldef\tempurl%
\url{https://doi.org/10.1109/ICWS.2004.1314784}
\showDOI{\tempurl}


\bibitem[\protect\citeauthoryear{Fente, Knutson, and Schexnayder}{Fente
  et~al\mbox{.}}{1999}]%
        {Fente1999betadist}
\bibfield{author}{\bibinfo{person}{Javier Fente}, \bibinfo{person}{Kraig
  Knutson}, {and} \bibinfo{person}{Cliff Schexnayder}.}
  \bibinfo{year}{1999}\natexlab{}.
\newblock \showarticletitle{Defining a Beta Distribution Function for
  Construction Simulation}. In \bibinfo{booktitle}{\emph{Proceedings of the
  31st Conference on Winter Simulation: Simulation---a Bridge to the Future -
  Volume 2}} (Phoenix, Arizona, USA) \emph{(\bibinfo{series}{WSC '99})}.
  \bibinfo{publisher}{ACM}, \bibinfo{address}{New York, NY, USA},
  \bibinfo{pages}{1010--1015}.
\newblock
\showISBNx{0-7803-5780-9}
\urldef\tempurl%
\url{https://doi.org/10.1145/324898.324983}
\showDOI{\tempurl}


\bibitem[\protect\citeauthoryear{Fung}{Fung}{1995}]%
        {fung1995compiling}
\bibfield{author}{\bibinfo{person}{Pascale Fung}.}
  \bibinfo{year}{1995}\natexlab{}.
\newblock \showarticletitle{Compiling bilingual lexicon entries from a
  non-parallel English-Chinese corpus}. In
  \bibinfo{booktitle}{\emph{Proceedings of the Third Workshop on Very Large
  Corpora}}. \bibinfo{pages}{173--183}.
\newblock


\bibitem[\protect\citeauthoryear{Fung}{Fung}{1998}]%
        {Fung-98}
\bibfield{author}{\bibinfo{person}{Pascale Fung}.}
  \bibinfo{year}{1998}\natexlab{}.
\newblock \showarticletitle{A statistical view on bilingual lexicon extraction:
  from parallel corpora to non-parallel corpora}.
\newblock In \bibinfo{booktitle}{\emph{Machine Translation and the Information
  Soup}}. \bibinfo{publisher}{Springer}, \bibinfo{pages}{1--17}.
\newblock


\bibitem[\protect\citeauthoryear{Gupta and Nadarajah}{Gupta and
  Nadarajah}{2004}]%
        {gupta2004betadist}
\bibfield{author}{\bibinfo{person}{Arjun~K Gupta} {and}
  \bibinfo{person}{Saralees Nadarajah}.} \bibinfo{year}{2004}\natexlab{}.
\newblock \bibinfo{booktitle}{\emph{Handbook of beta distribution and its
  applications}}.
\newblock \bibinfo{publisher}{CRC press}.
\newblock


\bibitem[\protect\citeauthoryear{Holman, Brown, Wichmann, M{\"u}ller,
  Velupillai, Hammarstr{\"o}m, Sauppe, Jung, Bakker, Brown,
  et~al\mbox{.}}{Holman et~al\mbox{.}}{2011}]%
        {holman2011automated}
\bibfield{author}{\bibinfo{person}{Eric~W Holman}, \bibinfo{person}{Cecil~H
  Brown}, \bibinfo{person}{S{\o}ren Wichmann}, \bibinfo{person}{Andr{\'e}
  M{\"u}ller}, \bibinfo{person}{Viveka Velupillai}, \bibinfo{person}{Harald
  Hammarstr{\"o}m}, \bibinfo{person}{Sebastian Sauppe}, \bibinfo{person}{Hagen
  Jung}, \bibinfo{person}{Dik Bakker}, \bibinfo{person}{Pamela Brown},
  {et~al\mbox{.}}} \bibinfo{year}{2011}\natexlab{}.
\newblock \showarticletitle{Automated dating of the world's language families
  based on lexical similarity}.
\newblock \bibinfo{journal}{\emph{Current Anthropology}} \bibinfo{volume}{52},
  \bibinfo{number}{6} (\bibinfo{year}{2011}), \bibinfo{pages}{841--875}.
\newblock


\bibitem[\protect\citeauthoryear{Howard}{Howard}{1960}]%
        {howard1960dynamic}
\bibfield{author}{\bibinfo{person}{Ronald~A Howard}.}
  \bibinfo{year}{1960}\natexlab{}.
\newblock \bibinfo{booktitle}{\emph{Dynamic Programming and Markov Processes.}}
\newblock \bibinfo{publisher}{The M.I.T. Press}.
\newblock


\bibitem[\protect\citeauthoryear{Ishida}{Ishida}{2016}]%
        {ishida2016intercultural}
\bibfield{author}{\bibinfo{person}{Toru Ishida}.}
  \bibinfo{year}{2016}\natexlab{}.
\newblock \showarticletitle{Intercultural Collaboration and Support Systems: A
  Brief History}. In \bibinfo{booktitle}{\emph{International Conference on
  Principles and Practice of Multi-Agent Systems (PRIMA 2016)}}. Springer,
  \bibinfo{pages}{3--19}.
\newblock


\bibitem[\protect\citeauthoryear{{Ishida}, {Murakami}, {Lin}, {Nakaguchi}, and
  {Otani}}{{Ishida} et~al\mbox{.}}{2018}]%
        {ishida2018langrid}
\bibfield{author}{\bibinfo{person}{T. {Ishida}}, \bibinfo{person}{Y.
  {Murakami}}, \bibinfo{person}{D. {Lin}}, \bibinfo{person}{T. {Nakaguchi}},
  {and} \bibinfo{person}{M. {Otani}}.} \bibinfo{year}{2018}\natexlab{}.
\newblock \showarticletitle{Language Service Infrastructure on the Web: The
  Language Grid}.
\newblock \bibinfo{journal}{\emph{Computer}} \bibinfo{volume}{51},
  \bibinfo{number}{6} (\bibinfo{date}{June} \bibinfo{year}{2018}),
  \bibinfo{pages}{72--81}.
\newblock
\showISSN{0018-9162}
\urldef\tempurl%
\url{https://doi.org/10.1109/MC.2018.2701643}
\showDOI{\tempurl}


\bibitem[\protect\citeauthoryear{Lewis, Simons, and Fennig}{Lewis
  et~al\mbox{.}}{2015}]%
        {Lewis-15}
\bibfield{editor}{\bibinfo{person}{M.~Paul Lewis}, \bibinfo{person}{Gary~F.
  Simons}, {and} \bibinfo{person}{Charles~D. Fennig}} (Eds.).
  \bibinfo{year}{2015}\natexlab{}.
\newblock \bibinfo{booktitle}{\emph{Ethnologue: Languages of the World}
  (\bibinfo{edition}{18th} ed.)}.
\newblock \bibinfo{publisher}{SIL International}, \bibinfo{address}{Dallas,
  Texas}.
\newblock
\urldef\tempurl%
\url{http://www.ethnologue.com}
\showURL{%
\tempurl}


\bibitem[\protect\citeauthoryear{Murakami}{Murakami}{2019}]%
        {murakami2019indonesia}
\bibfield{author}{\bibinfo{person}{Yohei Murakami}.}
  \bibinfo{year}{2019}\natexlab{}.
\newblock \showarticletitle{Indonesia language sphere: an ecosystem for
  dictionary development for low-resource languages}. In
  \bibinfo{booktitle}{\emph{Journal of Physics: Conference Series}},
  Vol.~\bibinfo{volume}{1192}. IOP Publishing, \bibinfo{pages}{012001}.
\newblock


\bibitem[\protect\citeauthoryear{Nasution}{Nasution}{2018}]%
        {nasution2018PHMT}
\bibfield{author}{\bibinfo{person}{Arbi~Haza Nasution}.}
  \bibinfo{year}{2018}\natexlab{}.
\newblock \showarticletitle{Pivot-based Hybrid Machine Translation to Support
  Multilingual Communication for Closely Related Languages}.
\newblock \bibinfo{journal}{\emph{World Transactions on Engineering and
  Technology Education}} \bibinfo{volume}{16}, \bibinfo{number}{2}
  (\bibinfo{year}{2018}), \bibinfo{pages}{12--17}.
\newblock


\bibitem[\protect\citeauthoryear{Nasution, Kadir, Murakami, and
  Ishida}{Nasution et~al\mbox{.}}{2020}]%
        {nasution2020towardsplanning}
\bibfield{author}{\bibinfo{person}{Arbi~Haza Nasution},
  \bibinfo{person}{Evizal~Abdul Kadir}, \bibinfo{person}{Yohei Murakami}, {and}
  \bibinfo{person}{Toru Ishida}.} \bibinfo{year}{2020}\natexlab{}.
\newblock \bibinfo{booktitle}{\emph{Toward Formalization of Comprehensive
  Bilingual Dictionaries Creation Planning as Constraint Optimization
  Problem}}.
\newblock \bibinfo{publisher}{Springer Singapore},
  \bibinfo{address}{Singapore}, \bibinfo{pages}{41--54}.
\newblock
\showISBNx{978-981-15-2655-8}
\urldef\tempurl%
\url{https://doi.org/10.1007/978-981-15-2655-8_3}
\showDOI{\tempurl}


\bibitem[\protect\citeauthoryear{Nasution, Murakami, and Ishida}{Nasution
  et~al\mbox{.}}{2016}]%
        {nasution2016pivot}
\bibfield{author}{\bibinfo{person}{Arbi~Haza Nasution}, \bibinfo{person}{Yohei
  Murakami}, {and} \bibinfo{person}{Toru Ishida}.}
  \bibinfo{year}{2016}\natexlab{}.
\newblock \showarticletitle{Constraint-Based Bilingual Lexicon Induction for
  Closely Related Languages}. In \bibinfo{booktitle}{\emph{Proceedings of the
  Tenth International Conference on Language Resources and Evaluation (LREC
  2016)}} (Portorož, Slovenia, 23-28). \bibinfo{address}{Paris, France},
  \bibinfo{pages}{3291--3298}.
\newblock
\showISBNx{978-2-9517408-9-1}


\bibitem[\protect\citeauthoryear{Nasution, Murakami, and Ishida}{Nasution
  et~al\mbox{.}}{2017a}]%
        {nasution2017pivot}
\bibfield{author}{\bibinfo{person}{Arbi~Haza Nasution}, \bibinfo{person}{Yohei
  Murakami}, {and} \bibinfo{person}{Toru Ishida}.}
  \bibinfo{year}{2017}\natexlab{a}.
\newblock \showarticletitle{A Generalized Constraint Approach to Bilingual
  Dictionary Induction for Low-Resource Language Families}.
\newblock \bibinfo{journal}{\emph{ACM Trans. Asian Low-Resour. Lang. Inf.
  Process.}} \bibinfo{volume}{17}, \bibinfo{number}{2}, Article
  \bibinfo{articleno}{9} (\bibinfo{date}{Nov.} \bibinfo{year}{2017}),
  \bibinfo{numpages}{29}~pages.
\newblock
\showISSN{2375-4699}
\urldef\tempurl%
\url{https://doi.org/10.1145/3138815}
\showDOI{\tempurl}


\bibitem[\protect\citeauthoryear{Nasution, Murakami, and Ishida}{Nasution
  et~al\mbox{.}}{2017b}]%
        {nasution2017plan}
\bibfield{author}{\bibinfo{person}{Arbi~Haza Nasution}, \bibinfo{person}{Yohei
  Murakami}, {and} \bibinfo{person}{Toru Ishida}.}
  \bibinfo{year}{2017}\natexlab{b}.
\newblock \showarticletitle{Plan Optimization for Creating Bilingual
  Dictionaries of Low-Resource Languages}. In \bibinfo{booktitle}{\emph{2017
  International Conference on Culture and Computing (Culture and Computing)}}.
  \bibinfo{pages}{35--41}.
\newblock
\urldef\tempurl%
\url{https://doi.org/10.1109/Culture.and.Computing.2017.21}
\showDOI{\tempurl}


\bibitem[\protect\citeauthoryear{Nasution, Murakami, and Ishida}{Nasution
  et~al\mbox{.}}{2018}]%
        {nasution2018collab}
\bibfield{author}{\bibinfo{person}{Arbi~Haza Nasution}, \bibinfo{person}{Yohei
  Murakami}, {and} \bibinfo{person}{Toru Ishida}.}
  \bibinfo{year}{2018}\natexlab{}.
\newblock \showarticletitle{Designing a Collaborative Process to Create
  Bilingual Dictionaries of Indonesian Ethnic Languages}. In
  \bibinfo{booktitle}{\emph{Proceedings of the Eleventh International
  Conference on Language Resources and Evaluation (LREC 2018)}} (Miyazaki,
  Japan, 7-12). \bibinfo{publisher}{European Language Resources Association
  (ELRA)}, \bibinfo{address}{Paris, France}, \bibinfo{pages}{3397--3404}.
\newblock
\showISBNx{979-10-95546-00-9}


\bibitem[\protect\citeauthoryear{Nasution, Murakami, and Ishida}{Nasution
  et~al\mbox{.}}{2019}]%
        {nasution2019simcluster}
\bibfield{author}{\bibinfo{person}{Arbi~Haza Nasution}, \bibinfo{person}{Yohei
  Murakami}, {and} \bibinfo{person}{Toru Ishida}.}
  \bibinfo{year}{2019}\natexlab{}.
\newblock \showarticletitle{Generating Similarity Cluster of Indonesian
  Languages with Semi-Supervised Clustering}.
\newblock \bibinfo{journal}{\emph{International Journal of Electrical and
  Computer Engineering (IJECE)}} \bibinfo{volume}{9}, \bibinfo{number}{1}
  (\bibinfo{year}{2019}), \bibinfo{pages}{1--8}.
\newblock


\bibitem[\protect\citeauthoryear{Nasution, Syafitri, Setiawan, and
  Suryani}{Nasution et~al\mbox{.}}{2017c}]%
        {nasution2017PHMT}
\bibfield{author}{\bibinfo{person}{Arbi~Haza Nasution}, \bibinfo{person}{Nesi
  Syafitri}, \bibinfo{person}{Panji~Rahmat Setiawan}, {and}
  \bibinfo{person}{Des Suryani}.} \bibinfo{year}{2017}\natexlab{c}.
\newblock \showarticletitle{Pivot-Based Hybrid Machine Translation to Support
  Multilingual Communication}. In \bibinfo{booktitle}{\emph{2017 International
  Conference on Culture and Computing (Culture and Computing)}}.
  \bibinfo{pages}{147--148}.
\newblock
\urldef\tempurl%
\url{https://doi.org/10.1109/Culture.and.Computing.2017.22}
\showDOI{\tempurl}


\bibitem[\protect\citeauthoryear{Rapp}{Rapp}{1995}]%
        {rapp1995identifying}
\bibfield{author}{\bibinfo{person}{Reinhard Rapp}.}
  \bibinfo{year}{1995}\natexlab{}.
\newblock \showarticletitle{Identifying word translations in non-parallel
  texts}. In \bibinfo{booktitle}{\emph{Proceedings of the 33rd annual meeting
  on Association for Computational Linguistics}}. Association for Computational
  Linguistics, \bibinfo{pages}{320--322}.
\newblock


\bibitem[\protect\citeauthoryear{Russell and Norvig}{Russell and
  Norvig}{2016}]%
        {russell2016artificial}
\bibfield{author}{\bibinfo{person}{Stuart~J Russell} {and}
  \bibinfo{person}{Peter Norvig}.} \bibinfo{year}{2016}\natexlab{}.
\newblock \bibinfo{booktitle}{\emph{Artificial intelligence: a modern
  approach}}.
\newblock \bibinfo{publisher}{Malaysia; Pearson Education Limited,}.
\newblock


\bibitem[\protect\citeauthoryear{Soderland, Etzioni, Weld, Skinner, Bilmes,
  et~al\mbox{.}}{Soderland et~al\mbox{.}}{2009}]%
        {Soderland-09}
\bibfield{author}{\bibinfo{person}{Stephen Soderland}, \bibinfo{person}{Oren
  Etzioni}, \bibinfo{person}{Daniel~S Weld}, \bibinfo{person}{Michael Skinner},
  \bibinfo{person}{Jeff Bilmes}, {et~al\mbox{.}}}
  \bibinfo{year}{2009}\natexlab{}.
\newblock \showarticletitle{Compiling a massive, multilingual dictionary via
  probabilistic inference}. In \bibinfo{booktitle}{\emph{Proceedings of the
  Joint Conference of the 47th Annual Meeting of the ACL and the 4th
  International Joint Conference on Natural Language Processing of the AFNLP:
  Volume 1-Volume 1}}. Association for Computational Linguistics,
  \bibinfo{pages}{262--270}.
\newblock


\bibitem[\protect\citeauthoryear{Swadesh}{Swadesh}{1955}]%
        {swadesh1955towards}
\bibfield{author}{\bibinfo{person}{Morris Swadesh}.}
  \bibinfo{year}{1955}\natexlab{}.
\newblock \showarticletitle{Towards greater accuracy in lexicostatistic
  dating}.
\newblock \bibinfo{journal}{\emph{International journal of American
  linguistics}} \bibinfo{volume}{21}, \bibinfo{number}{2}
  (\bibinfo{year}{1955}), \bibinfo{pages}{121--137}.
\newblock


\bibitem[\protect\citeauthoryear{Tanaka and Umemura}{Tanaka and
  Umemura}{1994}]%
        {tanaka-94}
\bibfield{author}{\bibinfo{person}{Kumiko Tanaka} {and} \bibinfo{person}{Kyoji
  Umemura}.} \bibinfo{year}{1994}\natexlab{}.
\newblock \showarticletitle{Construction of a bilingual dictionary
  intermediated by a third language}. In \bibinfo{booktitle}{\emph{Proceedings
  of the 15th conference on Computational linguistics-Volume 1}}. Association
  for Computational Linguistics, \bibinfo{pages}{297--303}.
\newblock


\bibitem[\protect\citeauthoryear{White}{White}{1993}]%
        {white1993survey}
\bibfield{author}{\bibinfo{person}{Douglas~J White}.}
  \bibinfo{year}{1993}\natexlab{}.
\newblock \showarticletitle{A survey of applications of Markov decision
  processes}.
\newblock \bibinfo{journal}{\emph{Journal of the Operational Research Society}}
  \bibinfo{volume}{44}, \bibinfo{number}{11} (\bibinfo{year}{1993}),
  \bibinfo{pages}{1073--1096}.
\newblock


\bibitem[\protect\citeauthoryear{Wushouer, Lin, Ishida, and Hirayama}{Wushouer
  et~al\mbox{.}}{2015}]%
        {wushouer-15}
\bibfield{author}{\bibinfo{person}{Mairidan Wushouer}, \bibinfo{person}{Donghui
  Lin}, \bibinfo{person}{Toru Ishida}, {and} \bibinfo{person}{Katsutoshi
  Hirayama}.} \bibinfo{year}{2015}\natexlab{}.
\newblock \showarticletitle{A Constraint Approach to Pivot-Based Bilingual
  Dictionary Induction}.
\newblock \bibinfo{journal}{\emph{ACM Trans. Asian Low-Resour. Lang. Inf.
  Process.}} \bibinfo{volume}{15}, \bibinfo{number}{1}, Article
  \bibinfo{articleno}{4} (\bibinfo{date}{Nov.} \bibinfo{year}{2015}),
  \bibinfo{numpages}{26}~pages.
\newblock
\showISSN{2375-4699}
\urldef\tempurl%
\url{https://doi.org/10.1145/2723144}
\showDOI{\tempurl}


\bibitem[\protect\citeauthoryear{Yu, Buyya, and Tham}{Yu et~al\mbox{.}}{2005}]%
        {yu2005workflowgrid}
\bibfield{author}{\bibinfo{person}{Jia Yu}, \bibinfo{person}{R. Buyya}, {and}
  \bibinfo{person}{Chen~Khong Tham}.} \bibinfo{year}{2005}\natexlab{}.
\newblock \showarticletitle{Cost-based scheduling of scientific workflow
  applications on utility grids}. In \bibinfo{booktitle}{\emph{First
  International Conference on e-Science and Grid Computing (e-Science'05)}}.
  \bibinfo{pages}{8 pp.--147}.
\newblock
\urldef\tempurl%
\url{https://doi.org/10.1109/E-SCIENCE.2005.26}
\showDOI{\tempurl}


\end{thebibliography}

\appendix

\end{document}